\documentclass[twoside]{article}

\usepackage[accepted]{aistats2023}
\usepackage[round]{natbib}

\usepackage{amsfonts, amssymb}
\usepackage{mathrsfs}
\usepackage{graphicx}
\usepackage{cancel}
\usepackage{booktabs}
\usepackage{algorithm}
\usepackage{algorithmic}
\usepackage{multirow}
\usepackage[backref=page]{hyperref}
\usepackage{cleveref}
\usepackage{caption}
\usepackage{subcaption}
\usepackage{textcomp}

\DeclareMathOperator*{\argmin}{\arg\,\min}
\newcommand{\Qcal}{\mathcal{Q}}

\newcommand{\R}{\mathbb{R}}

\newcommand{\E}{\mathbb{E}}
\newcommand{\DKL}{\mathbb{D}}
\renewcommand{\d}{\,\mathrm{d}}
\newcommand{\inv}{^{-1}}
\newcommand{\diag}{\operatorname{diag}}
\usepackage{tikz}
\newcommand*\circled[1]{\tikz[baseline=(char.base)]{
            \node[shape=circle,draw,inner sep=2pt] (char) {#1};}}
\usepackage[normalem]{ulem}

\usepackage{amsthm}
\newtheorem{lemma}{Lemma}

\renewcommand{\textuparrow}{$\uparrow$}
\renewcommand{\textdownarrow}{$\downarrow$}

\crefname{section}{Sec.}{Sections}
\crefname{appendix}{App.}{Appendices}
\crefname{algorithm}{Alg.}{Algorithms}
\crefname{equation}{Eq.}{Eqs.}
\crefname{figure}{Fig.}{Figures}
\creflabelformat{equation}{#2\textup{#1}#3} 

\begin{document}

%

%

\twocolumn[

\aistatstitle{The Lie-Group Bayesian Learning Rule}

\aistatsauthor{ Eren Mehmet K{\i}ral \And Thomas M\"ollenhoff \And Mohammad Emtiyaz Khan }

\aistatsaddress{ RIKEN Center for AI Project \And RIKEN Center for AI Project \And RIKEN Center for AI projet } ]

\begin{abstract}
   The Bayesian Learning Rule provides a framework for generic algorithm design but can be difficult to use for three reasons.~First, it requires a specific parameterization of exponential family.~Second, it uses gradients which can be difficult to compute.~Third, its update may not always stay on the manifold.~We address these difficulties by proposing an extension based on Lie-groups where posteriors are parametrized through transformations of an arbitrary base distribution and updated via
   the group's exponential map.~This simplifies all three difficulties for many cases, providing flexible parametrizations through group's action, simple gradient computation through reparameterization, and updates that always stay on the manifold.~We use the new learning rule to derive a new algorithm for deep learning  with desirable biologically-plausible attributes to learn sparse features.~Our work opens a new frontier for the design of new algorithms by exploiting Lie-group structures.
 \end{abstract}

\section{INTRODUCTION}
\label{sec:intro}

The recently proposed Bayesian Learning Rule (BLR) of \citet{KhRu21} provides a general framework to derive many well-known algorithms from fields such as optimization, deep learning, and graphical models.~The rule uses natural-gradient descent to find approximations of the generalized posterior distribution and can recover both Bayesian and non-Bayesian algorithms by employing various exponential-family (EF) distributions.~It has been used to design new algorithms, for instance,
for uncertainty estimation in deep learning \citep{KhNi18, OsSw19, LiKh19b, MeBa20, MoKh23}.~Any improvements to the BLR framework can potentially be useful for such algorithm design as well.

Despite its usefulness, the BLR can be difficult to use for three reasons.~First, it relies heavily on pairings of natural and expectation parameters of the EFs which do not naturally exist for generic distributions and can make it difficult to apply the BLR to such cases~\citep{LiKh19b}.
Second, the BLR requires natural-gradients whose computation is not always straightforward and requires tricks that need to be invented for each specific case, for example, \citet{LiKh19} use Stein's identity for Gaussians and~\citet{MeBa20} use Gumbel-softmax trick for Bernoulli distributions.~A last difficulty is that the BLR updates are not always guaranteed to stay within the manifold of distributions, which may require additional modifications \citep{LiSc20}.
Our goal here is to address these three difficulties with the BLR.

\begin{figure*}[t!]
    \centering
  \begin{subfigure}[b]{0.33\textwidth}
    \centering
 	\includegraphics[height=3.6cm]{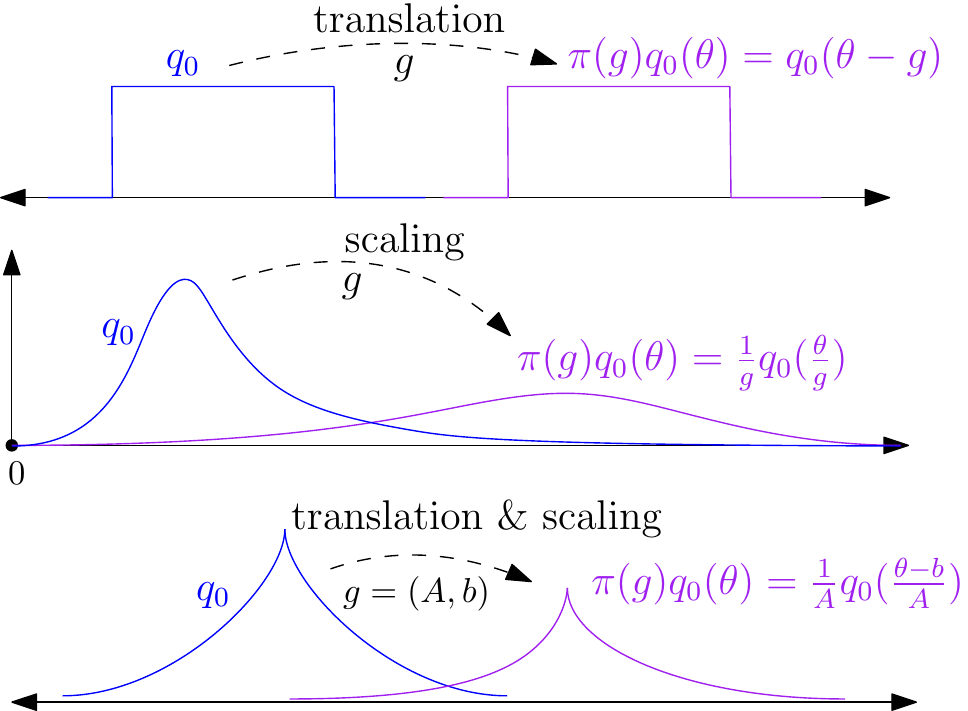}
 	\caption{}\label{fig:subfigBLR}
 	\end{subfigure}
 \begin{subfigure}[b]{0.33\textwidth}
    \centering
 	\includegraphics[height=3.6cm]{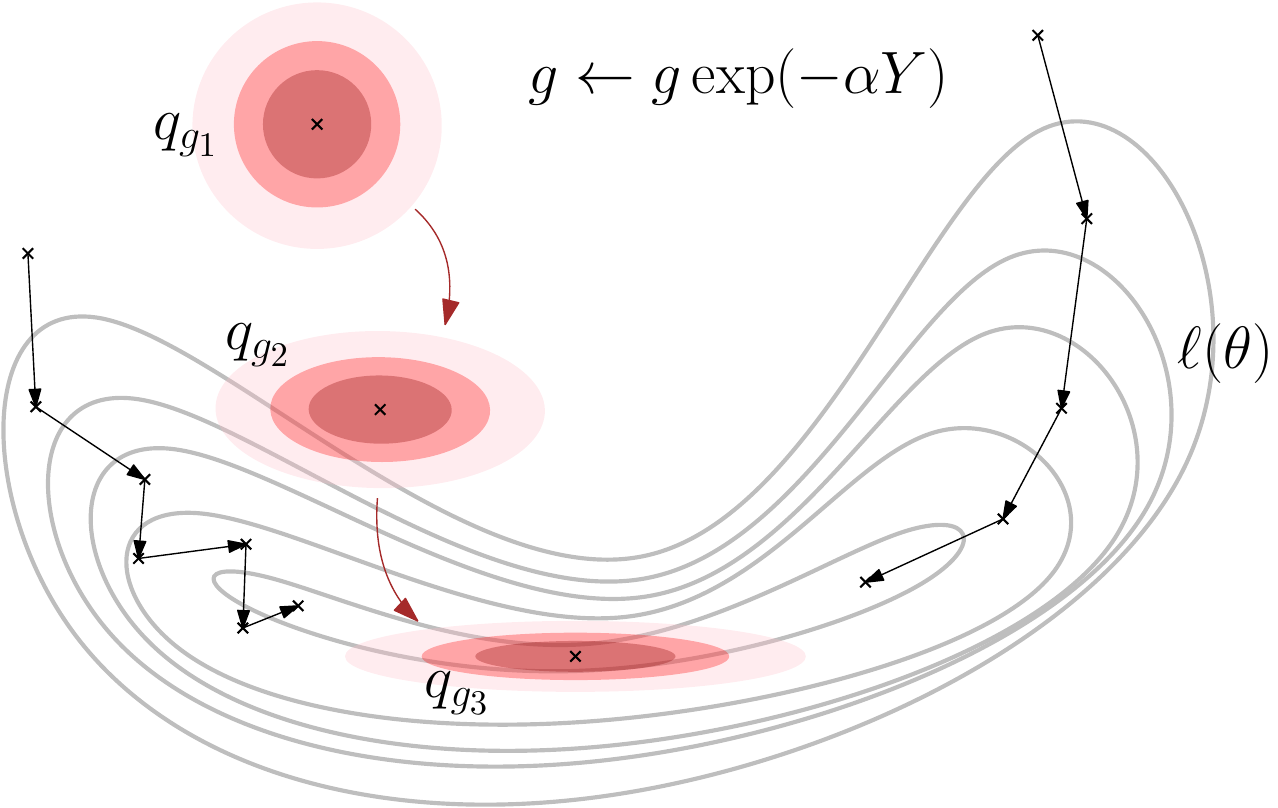}
 	\caption{}\label{fig:updates}
 \end{subfigure}
 	\begin{subfigure}[b]{0.33\textwidth}
    \centering
 	\includegraphics[height=3.6cm]{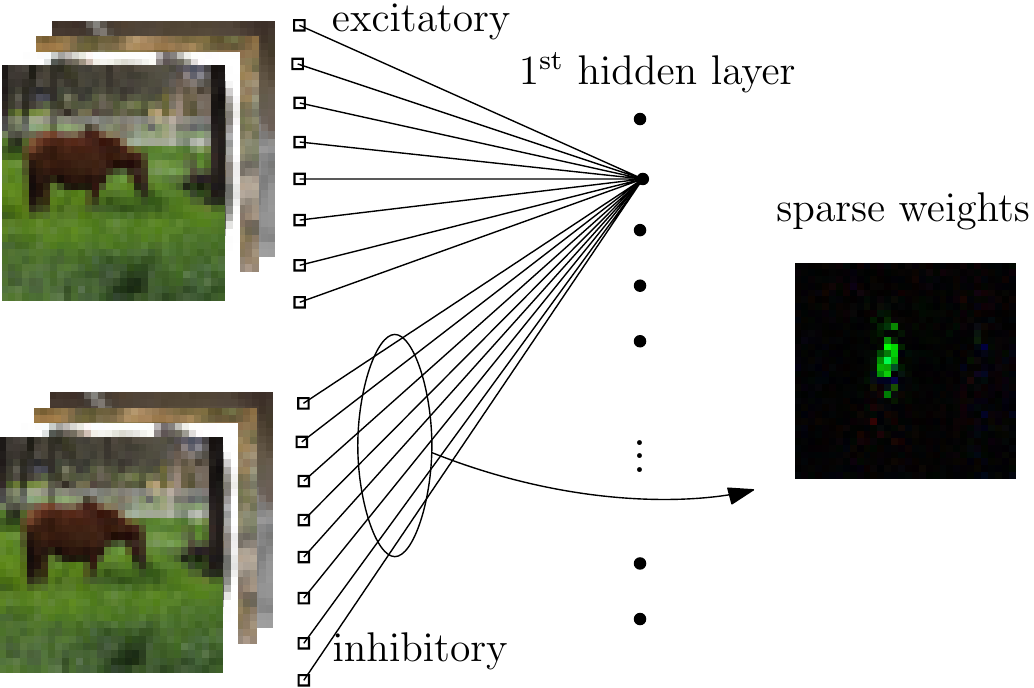}
 	\caption{}\label{fig:biologicalNN}
      \end{subfigure}
   \caption{(a) Posterior candidates are parametrized by transforming the base distribution $q_0$ over parameter $\theta$. The three figures show the additive, multiplicative, and affine groups, respectively. In each figure, a group element $g$ is applied to $q_0$ through the action $\pi(g)$, giving rise to a new candidate. The base distribution is set to uniform (top), Rayleigh (middle), and Laplace (bottom) respectively. (b) The Lie-group BLR uses the group's exponential map to update
   $g$ giving rise to new candidates $q_g= \pi(g) q_0$~(shown with red ovals); an exact update is given in~\Cref{eq:updateRuleFleshedOut}. The algorithm is different from standard gradient methods used in deep learning to learn $\theta$ (black arrows).
   (c) The Lie-group BLR with multiplicative group gives rise to a new algorithm to train neural networks with constraints similar to those found in biological neural networks. Specifically, the signs of the weights from a node are fixed, making the node either excitatory or inhibitory. This gives rise to sparse and localized features similar to those found in the receptive fields of the mammalian visual cortex.}  
 \end{figure*}

We propose an extension of the BLR based on Lie-groups where posterior candidates are parametrized through transformations of an arbitrary base distribution by using the group's \emph{action} on the model parameters.~For example, the additive group, denoted by $(\R,+)$, translates real scalar parameters by addition, and the multiplicative group, denoted by $(\R_{>0}, \times)$, scales positive scalar parameters by multiplication (\Cref{fig:subfigBLR}). Many popular distributions can be
parameterized this way, including both EF and non-EF distributions \citep{BaNi82, barndorff2012decomposition}.

We derive a new learning rule called the Lie-group BLR that uses the group's \emph{exponential map} to update candidate distributions (\Cref{fig:updates}). A linear approximation of the map coincides with the BLR for some distributions, but the new rule is much easier to use in many cases.~First, it does not depend on the EF parameterization but on the Lie-group action which is relatively easier to work with, for example, when using non-EF distributions such as the Laplace distribution.~Second, gradient computations are simplified by a
simple change of variables to push the derivative to the loss function, giving rise to a general yet easy-to-implement reparameterization trick.~Third, due to the closure property of the group, the update naturally stays within the manifold; no additional effort or approximations are required. The new rule also simplifies the computation of the Fisher matrix and inclusion of momentum. Overall, the new learning rule is much easier to use than the BLR.

We show three use-cases for algorithm design in deep learning by employing the additive, multiplicative, and affine groups respectively.  
The additive and affine groups result in algorithms similar to those used in deep learning, but the multiplicative group gives rise to a new kind of algorithm to train neural networks with biologically-plausible attributes.
We consider networks with nodes that are forced to be either excitatory or inhibitory by fixing the signs of their weights (\Cref{fig:biologicalNN}). This aims to mimic constraints such as those observed in the receptive fields of mammalian visual cortex~\citep{hubel1962receptive,olshausen1996emergence}. By design, the new algorithm preserves the signs of the weights by keeping each update within the manifold and ends up learning sparse and localized features
(\Cref{fig:biologicalNN}). The use case shows the usefulness of the new learning rule in designing algorithms that encourage
explainability, compositionality, and disentanglement~\citep{BeZh20, WhDo22}. 

\section{THE BAYESIAN LEARNING RULE}
\label{sec:bkg}

Given a loss function $\ell(\theta)$ over a model with parameter $\theta \in \Theta$, the BLR aims to find 
\begin{equation}\label{eq:BLProblem}
	q_* \in \argmin_{q\in \Qcal} ~ \E_{q} [\ell] - \tau\mathcal{H}(q), 
\end{equation}
where $q$ is a posterior candidate, $\Qcal$ is a space of candidate distributions, $\mathcal{H}(q)= - \int_{\Theta} q(\theta) \log q(\theta) \d \theta$ is the differential Shannon entropy, and $\tau > 0$ is a scalar  parameter, sometimes referred to as the temperature.
The first term favors regions with low losses, while the second term favors higher spread of $q$, and balancing them requires an exploration-exploitation tradeoff, favoring flatter regions of low loss.
The problem can also rewritten as an inference problem where we seek the best possible posterior candidate in $\mathcal{Q}$ by minimizing the Kullback-Leibler divergence,
\[
	\mathcal{E}(q) = \DKL(q\|p_\tau)
\]
where $p_\tau(\theta) \propto e^{-\frac{1}{\tau}\ell(\theta)}$ is the Gibbs posterior, sometimes referred to as the generalized posterior \citep{Ca07}.
When the loss corresponds to the log-joint distribution of a Bayesian model, $\Qcal$ is set to the space of all distribution and $\tau=1$, the solution in \eqref{eq:BLProblem} coincides with the posterior distribution; see \citet{Ze88}. When using $\tau\ne 1$, it is common for such cases to not scale the prior; see \citet[Eq. (2)]{OsSw19}.
Another interpretation is as a stochastic relaxation where the temperature $\tau$ is used to search for suitable minima \citep{GeGe84}. Such principles are commonly used in random search \citep{baba1981convergence}, stochastic optimization
\citep{spall2005introduction}, evolutionary strategies \citep{beyer2001theory}, global-optimization methods \citep{LeHe08}, and reinforcement learning \citep{williams1991function, mnih2016asynchronous}.

The BLR is a natural-gradient descent (NGD) algorithm to solve \eqref{eq:BLProblem}, and \citet{KhRu21} show that it can recover well-known algorithms from a variety of fields.
Specifically, they use \emph{minimal} exponential-family (EF) distributions of the form $q(\theta) \propto \gamma(\theta) e^{\left\langle \lambda, \phi(\theta) \right\rangle}$ parameterized by its natural parameter $\lambda$, where $\gamma(\theta)$ is a base measure, $\phi(\theta)$ is a sufficient statistics, and $\langle \cdot, \cdot \rangle$ is an inner product.
The BLR solves \eqref{eq:BLProblem} by updating $\lambda$ as follows, 
\begin{equation}
   \lambda \leftarrow \lambda - \alpha \frac{\partial}{\partial \mu} \left[ \E_q [\ell] - \tau \mathcal{H}(q) \right]
   \label{eq:BLR}
\end{equation}
where $\alpha >0$ is the learning rate and the gradients are taken with respect to the expectation parameter $\mu = \E_q[ \phi ]$. The BLR can recover many existing algorithms as special cases by simply changing the EF form and employing additional approximations to the gradient. \citet{KhRu21} show this by deriving gradient descent, Newton's method, and several deep-learning optimizers such as RMSprop and Adam, as well as message passing
algorithms,
such as, Kalman filters. Design of new algorithms is also possible, for example, for Bayesian deep learning \citet{KhNi18, OsSw19, LiKh19b, MeBa20, MoKh23}.

Despite its usefulness, the BLR can be difficult to use in many cases. First, the BLR update makes use of the pair $(\lambda, \mu)$, which makes its application difficult for other distributions where such pair is not available.
For example, for mixture of EFs, such pairs do not naturally exist, and special restrictions on the distribution are required to derive BLR-like updates; see \citet{LiKh19}. Some headway has been made for curved EFs too, for example, \ \citet{LiNi21} propose a local parametrization of structured Gaussian covariances, but deriving BLR-style updates for generic distributions remains an open problem.

Second, the gradient with respect to $\mu$ is not always straightforward to compute. For Gaussians, we can do this easily by using Stein's identity \citep{LiKh19} which reduces the computation to that of $\E_q[\nabla_\theta \ell(\theta)]$ and $\E_q[\nabla_\theta^2 \ell(\theta)]$ \citep[Eqs. 10-11]{KhRu21}. However, this trick does not generalize to arbitrary distributions. One option is to compute separately the Fisher and gradient with respect to $\lambda$ \citep[App. F]{KhLi17} but this does not work well
in practice due to large size of the Fisher matrix and also numerical difficulties arising due to noisy Fisher when estimated using minibatches; see \cite{SaKn13}. Third, $\lambda$ obtained by \eqref{eq:BLR} may not always be valid natural parameters, that is, the steps might go outside the EF manifold. \citet[App. D1]{KhNi18} discuss this problem for Gaussians where the update may result in negative variances. The problem is solved in \citet{LiSc20} by using Riemannian
gradient descent but such solutions need to be custom designed for specific cases which is tedious and cumbersome.

\section{THE LIE-GROUP BAYESIAN LEARNING RULE}

In this paper, we address the difficulties of the BLR described in the previous section by proposing a Lie-group based extension of the BLR. We start by describing Lie groups and their actions, followed by parameterization and exponential map, and finish the section by deriving the new learning rule. Readers unfamiliar with Lie groups can refer to~\citet[Chapter 7, 8, and 20]{Le13} for a detailed study.

\subsection{Lie groups and their actions}

We denote by $(G,*)$ a Lie-group, where $G$ is a set with a binary operation $*$ satisfying the properties of associativity, existence of an identity element and inverses. These mean three things: first, $g*(h*k) = (g*h)*k$ for all $g,h,k \in G$; second, there exists an identity element $e \in G$ such that $e*g = g*e = g$ for all $g\in G$; finally, for any $g\in G$ there exists an inverse element which we denote by $g\inv \in G$ such that $g*g\inv = g\inv * g = e$.
The Lie-group is a group, a smooth manifold, and both of its binary group-operation and inversion are smooth. 
For groups written with a multiplicative notation, it is common to write $gh$ in place of $g*h$.
A smooth manifold is locally diffeomorphic to Euclidean space, that is, there are infinitely differentiable invertible mappings between local patches of $G$ and $\R^m$, where $m$ is called the dimension of the manifold.

As an example, consider $(\R,+)$ where $\R$ is a 1-dimensional smooth manifold and, together with addition, it makes a Lie group. The identity element is $0$ and inverse of a given element $x \in \R$ is written as $-x$. Another example is the set of positive reals with multiplication, which forms the group~$(\R_{>0}, \times)$. 
A useful property is that if $G_1$ and $G_2$ are two Lie groups, then their Cartesian product $G_1 \times G_2$ is also a Lie group. The definition is extended to vector by repeating it $P$ times to get Lie-groups $(\R^P, +)$ and $(\R_{>0}^P, \times)$, where the addition and multiplication are both applied component-wise.

Given a manifold $\Theta$ of parameters $\theta$, we can define the \emph{action} of the Lie group on $\Theta$. The action is a smooth map  
$G \times \Theta \rightarrow \Theta$ mapping every $g\in G$ and $\theta\in\Theta$ as $(g, \theta) \mapsto g\cdot \theta$, where `$\cdot$' denotes an operation satisfying $(gh)\cdot \theta = g\cdot(h\cdot \theta)$ and $e\cdot \theta = \theta$. 
As an example, consider the group $(\R^P, +)$ and parameter-space $\theta \in \R^P$, then the action is the map $g\cdot\theta = g + \theta$. Here, both $G$ and $\Theta$ are the same space $\R^P$, but they can also be different. For example, consider the affine group where $G = \operatorname{Aff}_P(\R)$, consisting of pairs $(A,b)$ with $A$ an invertible matrix of size
$P\times P$ and $b \in \R^P$, and the group operation given by $(A_1,b_1)(A_2,b_2) = (A_1A_2, A_1b_2 + b_1)$. Say $\Theta = \R^P$, same as before. The action of $G$ on $\Theta$ is then given as $(A,b)\cdot \theta = A\theta + b$, for any $g = (A,b)\in G$ and $\theta \in \Theta$.

\subsection{Lie group parametrization}

Using the action of $G$ on $\Theta$, we can define another action on the space of measures by pushforwards. To be precise, given a measure $\nu$ on $\Theta$ and a measurable set $A\subseteq \Theta$ we define $\pi(g) \nu (A)= \nu(g\inv A)$ where $gA = \{g\cdot \theta: \theta \in A\}$. Considering probability measures of the form $\nu = q(\theta)\d\theta$, in terms of the probability density functions, we have
\begin{equation}\label{eq:groupActionOnDensity}
	(\pi(g) q)(\theta) = \left|\frac{\d(g\cdot \theta)}{\d \theta}\right|\inv q(g\inv \cdot \theta) ,
\end{equation}  
where $\d(g\cdot \theta)/\d\theta$ is the Jacobian determinant of $\theta \mapsto g \cdot \theta$.

We take a base distribution given with positive density $q_0$, and let the space of candidate distribution $\Qcal$ be the orbit of $q_0$ under the action of $G$, defined below,
\begin{equation}
  \Qcal = \{\pi(g) (q_0(\theta) \d \theta) : g \in G \}.
  \label{eq:qcal}
\end{equation}
This gives us a transitive action of $G$ on $\Qcal$. Also every $q \in \Qcal$ can be parametrized by group elements $g$ to write $q = q_g = \pi(g) q_0$. We will denote this parametrization by $\varphi: G\to \Qcal$ with $\varphi(g)= q_g$.

Here is an example. Take $q_0(\theta) = \theta e^{-\frac{1}{2}\theta^2}$, which is a parameter-free distribution on $\Theta = \R_{>0}$. The group $(\R_{>0}, \times)$ acts on $\Theta$ by  $g \cdot \theta = g\theta$, and the Jacobian of this map is simply $g$. The pushforward action of $g\in \R_{>0}$ on $q_0$ traverses the set of Rayleigh distributions
\begin{equation}\label{eq:rayleigh}
	\Qcal = \left\{q_g(\theta) = \tfrac{\theta}{g^2}e^{-\frac{\theta^2}{2 g^2}} : g \in \R_{>0}\right\},
\end{equation}
which is a family parametrized by the group.

The parameterization depends on the action over a group element $g$ and is different from those used for EF. The good news is that many EFs can be parameterized this way, for example, Gaussian and Bernoulli distribution. These are also sometimes referred to as the transformation families or models~\citep{barndorff2012decomposition}. The advantage of this parameterization is that it can be relatively easier to work with when using non-EF distributions such as the Laplace
distribution, which is useful to extend the BLR.

\subsection{The exponential map and Lie group updates}

Given the group parametrization  above, our goal is to find a group element $g_*$ such that
\[
	g_* \in \argmin_{g \in G} \, \mathcal{E}(q_g).
\]
We can find $g_*$ with an iterative update, for example, by slowly moving in the direction of fastest descent. This can be done by using the exponential map. 

Given a Lie group, its tangent space at identity, denoted by $T_e G$, is called the Lie algebra of~$G$; see the first figure in \Cref{fig:composition}. The exponential map is a smooth function `folding' the tangent space at identity to the group, which we denote by $\exp: T_eG \to G$. 
The map is well defined for all tangent vectors, and in fact it is one-to-one and onto in small neighborhoods around the $\mathbf{0}\in T_eG$ vector and $e\in G$. As an example, for matrix groups the exponential map is given by the Taylor series $\exp(X) = \sum_{n = 0}^\infty \frac{X^n}{n!}$. For diagonal matrices $X = \diag(\lambda_1, \ldots, \lambda_r)$ we can easily calculate $\exp(X) = \diag(e^{\lambda_1}, \ldots, e^{\lambda_r})$.

For any $X \in T_eG$, the exponential map defines paths $\gamma_X :\R \to G$ in $G$ via $\gamma_X(t) := \exp(tX)$ satisfying $\gamma_X(t_1)\gamma_X(t_2) = \gamma_X(t_1 + t_2)$. It is a path going through the identity at $t = 0$ in the direction of $X$, meaning $\gamma_X(0) = e$ and $\tfrac{\d}{\d t} \gamma_X(t)\big|_{t = 0} = X$.
At a particular $g \in G$, we can use an update of the form 
\[g \leftarrow g \exp(-\alpha X),\]
moving in the direction of $X\in T_e G$ by a step-size of $\alpha>0$. The direction can be the one with the fastest ascent. 

\subsection{Simplifying gradients through reparametrization}
\begin{figure*}[t]
\centering
\includegraphics[width=0.9\linewidth]{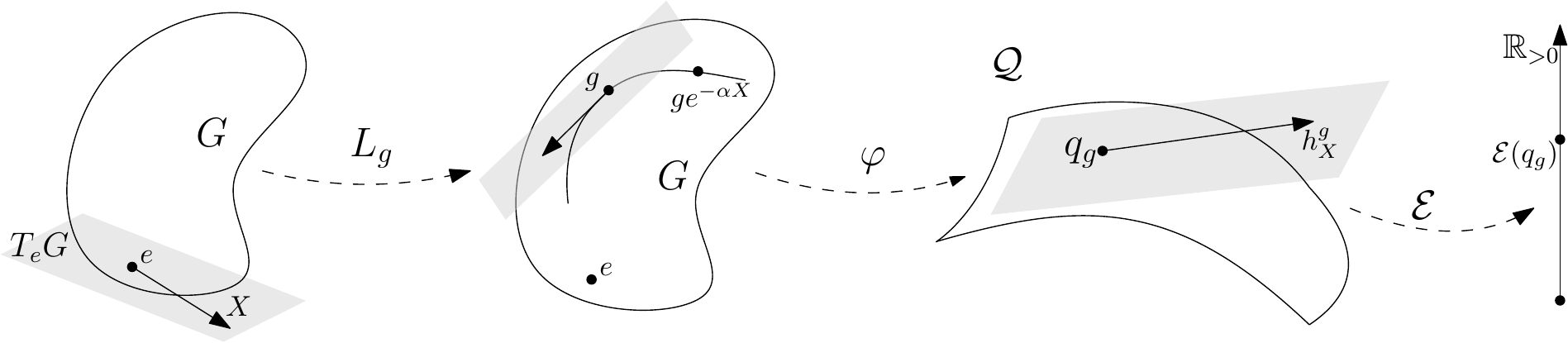}

\caption{$L_g$ is the multiplication-by-$g$ map and sends the identity element $e$ to $g$. Its differential takes the vector $X$ at the Lie algebra $T_eG$ to a vector tangent to $g$. The differential of $\varphi: g \mapsto q_g$ maps it to a tangent vector $h_X^g \in T_{q_g}\Qcal$. Among all such tangent directions we choose the  one corresponding to the direction of fastest ascent for the objective function $\mathcal{E}$. Then we update the $g$ using the exponential map, which defines a curve on $G$ through $g$ in a desired direction $-X$.}\label{fig:composition}
\end{figure*}

We will make use of the group's exponential map to derive a new learning rule. A summary of our approach is given in \Cref{fig:composition}, which relies on connecting tangent vectors $X\in T_e G$ to tangent vectors in $\mathcal{Q}$ which lie in the tangent space at a point $q_g$, denoted by $T_{q_g}\mathcal{Q}$. We start by showing the simplification of the gradient computation by using a change of variable to push the derivative to the loss function,
giving rise to a general yet easy-to-implement reparameterization trick.

We first parametrize $T_{q_g} \Qcal$ by the vectors in the Lie algebra and then compute the differential of $\mathcal{E}$ on vectors expressed in this way.
For every $g\in G$ the left multiplication
 map $L_g : G\to G$, defined as $L_g(h) =gh$, is an invertible smooth
 map on the manifold with its inverse being $L_{g\inv}$. These global diffeomorphisms give us linear maps $\mathrm{d} L_g (e) : T_eG \rightarrow T_g G$ between the vector spaces. The map $\mathrm{d} \varphi(g)$ then sends it to $T_{q_g}\Qcal$. Therefore call, $h_X^g = \d (\varphi \circ L_g) X \in T_{q_g}\Qcal$; see~\cref{fig:composition} for a visualization of these vectors and mappings. The perturbations of $\mathcal{E}$ at $q_g$ are given by the tangent vectors
 \[
 	h_X^g(\theta) = \left.\frac{\d}{\d t}  q_{g e^{tX}} (\theta) \right|_{t = 0}  \in T_{q_g}\Qcal.
 \]
 An explicit computation of these tangent vectors is given in \Cref{app:additive,app:multiplicative,app:affine} for certain Lie groups.
The linear map $X \mapsto h_X^g$ is surjective. Also, notice that the tangent vectors $h_X$ are  integrable functions on $\Theta$ satisfying $\int_\Theta h_X = 0$, since $\int_\Theta q_{ge^{tX}} (\theta) \d \theta$ is constant (equal to $1$) for all $t$. 
In what follows, we will drop $g$ from the notation $h_X^g$ whenever it is clear that we are working in the tangent space at $q_g$ or whenever $g$ does not matter.

We denote the differential of $\mathcal{E}$ at $q$ with the perturbation $h$ by $(\mathrm{d} \mathcal{E}(q))[h]$.
As shown in \cref{app:Fisher}, the differential in the direction $h_X$ at a point $q_g$ can be written as follows, 
\begin{equation}\label{eq:DEintegral}
	(\mathrm{d} \mathcal{E}(q_g)) [h_X] = \int h_X(\theta) \log \frac{{q_g}(\theta)}{p_\tau(\theta)} \d\theta,
\end{equation}
where $\mathrm{d} \mathcal{E}(q_g) : T_{q_g} \Qcal \to \R$ measures the change in $\mathcal{E}$ when $q_g$ is perturbed 
by variations $h_X$. The integral has two parts
\begin{align}\label{eq:DEsplit}
  \eqref{eq:DEintegral} =  \frac{1}{\tau}\underbrace{\int h_X(\theta) \ell(\theta)\d \theta}_{\circled{1}} 
  + \underbrace{\int h_X(\theta) \log q_{g}(\theta) \d \theta}_{\circled{2}}. 
\end{align}
Only the first part depends on the loss function $\ell$  and the second part is the differential of the entropy term in \eqref{eq:BLProblem}.

For $\circled{1}$, we start with the definition of $h_{X}(\theta)$ in the first equation below, and get the second and third equations by first plugging the definition of $q_g$ from \Cref{eq:qcal} and then changing variables $\theta \mapsto ge^{tX}\cdot \theta$, followed by a few more rearrangement afterwards, 
\begin{align}
 &\circled{1} = \frac{\d}{\d t} \left(\int q_{g e^{tX}}(\theta) \ell(\theta) \d \theta\right)\Bigg|_{t = 0} \notag\\
   & = \frac{\d}{\d t} \left(\int \frac{1}{\left|\frac{\d(ge^{tX}\cdot \theta)}{\d \theta} \right|} q_{0}((ge^{tX})\inv \cdot \theta) \ell(\theta) \d \theta\right)\Bigg|_{t = 0} \notag \\
 &= \frac{\d}{\d t} \left(\int q_0(\theta) \ell\left(g e^{tX} \cdot \theta\right) \d \theta \right)\Bigg|_{t = 0} \notag \\
 &= \int q_0(\theta) (\nabla_{\theta} \ell(g\cdot \theta) )^\top \frac{\d}{\d t} \left(ge^{tX} \cdot \theta \right) \bigg|_{t =0} \d \theta \notag\\
 &= \int q_g(\theta) (\nabla_{\theta} \ell( \theta) )^\top \left(\operatorname{Ad}_g(X) \cdot \theta\right) \d \theta. 
 \label{eq:circled1formula}
\end{align}
In the second-last equation, the derivative of $\ell$ appears because of the chain rule and, in the last line, we go back to $q_g$ by a change of variables $\theta \mapsto g\inv \cdot \theta$.
We denote
\[ \operatorname{Ad}_g(X) \cdot \theta = \frac{\d}{\d t} \left(ge^{tX} g^{-1} \cdot \theta \right) \big|_{t =0} \]
where $\operatorname{Ad}_g(X) = \frac{\d}{\d t} (g e^{tX} g\inv)\big|_{t = 0}$ maps from $T_eG$ to $T_eG$ and is called the adjoint representation of the Lie group. For commutative groups, the adjoint representation is the identity: $\operatorname{Ad}_g(X) = X$ for all $g, X$. 
The computation is simplified by using the change of variable and the derivative sits on the loss function which can be computed by automatic differentiation techniques. This is similar to pathwise gradient-estimators \citep{mohamed2020monte} but an advantage of using Lie-groups is that the path does need to be designed on a cases by case basis which is a major issue in designing generic reparameterization techniques
\citep{ruiz2016generalized, figurnov2018implicit}.

The entropic contribution is calculated similarly, using the same reparametrization technique (\cref{app:DEcalculation}). 
\begin{equation}\label{eq:circled2formula}
	\circled{2} = \int  \left(\nabla_{\theta} q_0(\theta)\right)^\top (X \cdot \theta) \d \theta,
\end{equation}
where we denote $X \cdot \theta = \frac{\d}{\d t} (e^{tX} \cdot \theta)\big|_{t =0}$. 
For the translation of these abstract formulas to particular cases, we refer the reader to \cref{sec:trinity}. Particular bases for $X \in T_eG$ are chosen and these integrals are calculated, giving us our concrete update rules.

\subsection{The new learning rule}

We are now ready to state our final rule. The Lie-Group BLR uses the following update
\begin{equation}\label{eq:updateRuleFleshedOut}
  g \leftarrow g\exp(-\alpha Y) \text{ where } h_Y = \big(\mathrm{d} \mathcal{E}(q_g)\big)^\sharp \in T_{q_g}\Qcal.
\end{equation}
Here, $(\mathrm{d} \mathcal{E}(q_g))^\sharp$ denotes the direction of fastest ascent at $q_g$, and $Y \in T_eG$ is such that its image $h_Y^g \in T_{q_g}\Qcal$ under $\d \varphi\circ \d L_g$ matches the direction of fastest ascent. Given such $Y$, the update naturally stay within the manifold due to the closure property of the group, where the exponential map folds the tangent vector back on the manifold. We will now explain the operator $\sharp$, also known as the musical-isomorphism sharp, and
its computation.

The operator $\sharp$ can be seen as the manifold analogue of the `transpose' of vectors required to define gradients in Euclidean spaces.
In Euclidean metric, the gradient $\nabla_\theta f$ of a function $f(\theta)$ is the direction of fastest ascent, 
and is also the transpose of the differential $\d f$ which is a row vector of referred to as the differential.
In general, if $f$ is a function that maps from a manifold $\mathcal{M}_1$ to $\mathcal{M}_2$, then its differential $\d f$ is a linear map between the tangent spaces $\mathrm{d} f(\theta): T_\theta \mathcal{M}_1 \longrightarrow T_{f(\theta)} \mathcal{M}_2$. If $\mathcal{M}_2 \subseteq \R$, then $\mathrm{d}f(\theta)$ is a linear functional taking tangent vectors to real numbers, also referred to as a cotangent vector $\mathrm{d}f(\theta) \in
T_\theta^*\mathcal{M}_1$, where $T_\theta^*\mathcal{M}_1$ is the dual of $T_\theta \mathcal{M}_1$.
The operator $\sharp$ is the Riemannian manifold analogue of the transpose.

We now give an exact characterization for a given metric. A metric, denoted by $\omega$, is a positive-definite, non-degenerate, symmetric, bilinear form on the tangent spaces of $\Qcal$. Fixing one of the variables in $\omega(\cdot,\cdot)$, we get a linear functional from the tangent space to reals, that is, a covector. Define, $\flat: T_q \Qcal \longrightarrow T_q^* \Qcal$, such that $(h_Y)^\flat = \omega(\cdot, h_Y)$, as a linear map called \emph{flat}. It is invertible because $\omega$ is non-degenerate. The inverse of this isomorphism is called \emph{sharp}, and is denoted by $\xi^\sharp \in T_q\Qcal$ for any~$\xi \in T^*_q\Qcal$. 
In vector notation, after choosing a basis $\{h_i\}_{i = 1, 2,\ldots, m}$ for $T_q\Qcal$ (where $m = \dim(\mathcal{Q})$), the metric is given by $\omega(v,w) = v^\top Fw$ for $F$ the symmetric matrix with entries $F_{ij} = \omega(h_i, h_j)$. Then $\flat$ maps a given (column) vector $v$ to the (row) vector $v^\top F$, and its inverse is $(w^\top)^\sharp = F\inv w.$ 

The differential $\mathrm{d}\mathcal{E}(q_g)$ of the functional $\mathcal{E}$ is a covector in $T_{q_g}^*\Qcal$. In vector notation, optimizing the linearization $\mathcal{E}(q) + (\mathrm{d}\mathcal{E}(q_g))[h_X]$ of the objective functional $\mathcal{E}$ near $q$, subject to the condition $\omega(h_X, h_X) \leq 1$ corresponds to  
\begin{align}
                      &\max_{v \in \R^m} ~ d^\top v ~~~ \text{ s.t. } ~~~ v^\top F v = 1,\notag\\
                      \text{where } &d = \begin{bmatrix} (\mathrm{d}\mathcal{E}(q_g))[h_1] & \cdots & (\mathrm{d}\mathcal{E}(q_g))[h_m]\end{bmatrix}^\top. \label{eq:DEinvectorform}
\end{align}
Solving it using Lagrange multipliers we get that $v$ must be a multiple of $F\inv d$, which equals $(\mathrm{d} \mathcal{E}(q_g))^\sharp$. Due to \eqref{eq:DEinvectorform}, this is the direction of fastest ascent with respect to the chosen metric $\omega$.

In the next section, we will give examples where the fastest direction can be obtained using $\nabla_\theta \ell$. An advantage of using Lie-groups is that the computation of the Fisher is simplified because it needs to be computed only once.~This is because, with the choice $h_X^g$ as tangent vectors, the metric depends only on $X$ and is independent of $g$. This is discussed in more detail in \Cref{app:Fisher}.
Using the Fisher metric is also natural because it arises as the second-order differential of our objective function $\mathcal{E}$, which means that the direction of fastest descent is aligned with minimizing the second-order approximation of $\mathcal{E}$.

It is also easy to include momentum in the Lie-group BLR. Because all $Y$ vectors are in $T_eG$, we can accumulate previous gradient steps and include momentum as follows,
	\begin{align*}
		M &\leftarrow (1-\beta) Y + \beta  M, \\
		g &\leftarrow g \exp(-\alpha M),
	\end{align*}
    where $\beta \in [0, 1)$, and the momentum term is initialized at $M = \mathbf{0}\in T_eG$ and $Y \in T_eG$ is found via \eqref{eq:updateRuleFleshedOut}. 
    
\section{NEW ALGORITHMS FOR DEEP LEARNING}
\label{sec:trinity}

We will now show three use-cases of the Lie-group BLR to design new algorithms for deep learning. In the BLR, new algorithms can be designed by changing the form of the EF. For the Lie-group BLR, we can do the same by employing various kinds of Lie-groups. The three examples we show will use the additive, multiplicative, and affine groups respectively. The additive and affine groups result in algorithms similar to those used in deep learning, but the multiplicative group gives rise to a new kind of algorithm to train neural networks with
biologically-plausible attributes.

We will also see that, in some cases, a linear approximation of the map coincides with the BLR; a summary of such results is given in \Cref{app:KhanRue}. The Lie-group BLR extends the BLR and provides yet another way to design new algorithms by employing various Lie-group.

\subsection{The additive group $G = (\R^P, +)$}
\label{sec:additive}

We start by specifying the Lie-group parameterization. We will assume $\Theta = \R^P$, then $G$ acts on $\Theta$ via $g\cdot \theta = \theta + g$. We set $q_0(\theta) = \prod_i \tilde{q}_0(\theta_i) \in \mathcal{P}^+(\Theta)$ where $\tilde{q}_0$ is a density function of an everywhere positive probability distribution on $\R$. The action of $G$ on the parameters induces an action on the probability distributions on $\Theta$ which in this case is given as $q_g(\theta) := (\pi(g)\cdot q_0)(\theta)
= q_0(\theta -g)$, and 
\[
	\Qcal := \{q_g(\theta)\d \theta : g \in \R^P \} = \{q_0(\theta - g) \d \theta: g \in \R^P\}.
\]

A detailed derivation of the Lie-group BLR is in \Cref{app:additive} which consists of 3 steps. First, we show that $\operatorname{Ad}_g(X) \cdot \theta = X \cdot \theta = X$ which is then plugged in \Cref{eq:circled1formula} to write the differential as $\E_{q_{g}}[\nabla_\theta \ell]$. Second, we show that the gradient of the entropy in \Cref{eq:circled2formula} is $\mathbf{0}$. Finally, the Fisher is an identity matrix multiplied by a constant. After these steps, \eqref{eq:updateRuleFleshedOut} reduces to the following, 
\begin{equation} \label{eq:additiveUpdate}
	g \leftarrow g - \alpha \E_{q_{g}}[\nabla_\theta \ell].
\end{equation}
This coincides with the update of~\citet[Eq. 7]{KhRu21} when $\mathcal{Q}$ is the set of Gaussians with variance 1. Clearly if $q_0$ is a standard Gaussian, then a translation will generate such Gaussians with the mean parameterized by $g$.

The Lie-group BLR generalizes the update obtained by \citet{KhRu21} to an arbitrary base distribution $q_0$. For this simple case, no linear approximation to the map is necessary to arrive at the BLR. This is because this group's exponential map is trivial, that is, already linear. We can also use distributions such as the uniform distribution, even though it is not an everywhere positive density distribution. Although the derivation does not allow for
choosing $q_0$ as a Dirac delta measure, such a choice will give us the classical gradient descent
$
	\theta \leftarrow \theta - \alpha \nabla_\theta \ell(\theta).
$

There is also a connection with anticorrelated noise injection~\citep{OrKe22}, which has been shown to perform better than gradient descent and its perturbed versions. Assume $q_0$ is centered around $\mathbf{0} \in \R^P$, so the mean of $q_{\theta}$ is $\theta$. If we use a single MC sample for the expected gradient, then the update rule is
\[
	\theta \leftarrow \theta - \alpha \nabla_\theta \ell(\theta + \xi),
\]
with noise $\xi \sim q_0$. This is exactly \citet[Eq.~3]{OrKe22} when the current iterate $\theta$ is set to the mean of $q$.

\begin{figure*}[t!]
\newdimen\figrasterwd
\figrasterwd\linewidth
  \parbox{\figrasterwd}{
\centering
    \parbox{.48\figrasterwd}{%
      \subcaptionbox{MNIST (MLP), additive, accuracy: $98.38$}{\includegraphics[width=\linewidth]{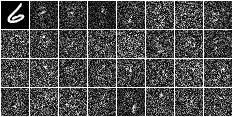}}
    }
    \hskip0.8em
    \parbox{.48\figrasterwd}{%
      \subcaptionbox{MNIST (MLP), multiplicative, accuracy: $98.59$}{\includegraphics[width=\linewidth]{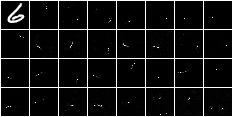}}
    }\\[0.25cm]
    \parbox{.48\figrasterwd}{%
      \subcaptionbox{CIFAR-10 (MLP), additive, accuracy: $58.85$}{\includegraphics[width=\linewidth]{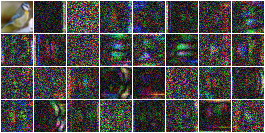}}
    }
    \hskip0.8em
    \parbox{.48\figrasterwd}{%
      \subcaptionbox{CIFAR-10 (MLP), multiplicative, accuracy: $59.19$}{\includegraphics[width=\linewidth]{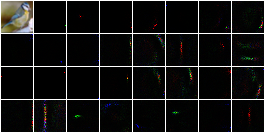}}
    }
  }
  \caption{Learning using multiplicative rule (\Cref{fig:multAlg}) on a network with biologically plausible attributes leads to sparse first-layer weights, which is in constrast to the additive rule (\Cref{fig:additiveAlg}). We visualize the weights connected to the neurons with highest activation under the input pattern in the top-left frame. The feature-detectors learned by our multiplicative rule show compositional and disentangled traits. This phenomenon occurs both for MNIST (a)--(b) and CIFAR-10 (c)--(d). }
  \label{fig:filters_add_and_mult}
\end{figure*}

\subsection{The multiplicative group $G = (\R_{>0}^P, \times)$}
\label{sec:multiplicative}

We consider networks with nodes that are forced to be either excitatory or inhibitory by fixing the signs of their weights (\Cref{fig:biologicalNN}). This aims to mimic constraints such as those observed in the receptive fields of mammalian visual cortex~\citep{hubel1962receptive,olshausen1996emergence}. We will use the multiplicative group to parameterize the distribution over the weights.

Let the parameter space be $\Theta = \R_{>0}^P$. For example consider the weights of a neural network whose signs are immutable and their magnitude is the only trainable parameter. In fact, we may assign certain nodes as excitatory (respectively inhibitory)---as is the case in biological neural circuitry---and set all the signs of weights for connections emanating from a cell as $+$ (respectively $-$).  This is also known as the $\pm$ trick, \citep{GhHa20}. This setup would then respect \emph{Dale's Law} from neurobiology, which is the assumption that a neuron has the same (excitatory or inhibitory) behaviour at all of its synapses, and that this does not change during training or stochastically. In neural networks this corresponds to keeping signs of weights fixed, see~\citet{AmWo1989}~and~\citet{BeZh21}.

$G$ acts on $\Theta$ by componentwise multiplication and given $q_0 \in \mathcal{P}^+(\Theta)$ as above, the transformations look like $q_g(\theta) := (\pi(g) q_0)(\theta) = \frac{1}{\prod_i g_i} q_0\left(\frac{\theta_1}{g_1}, \ldots, \frac{\theta_P}{g_P}\right).$
The manifold of candidate distributions is $\mathcal{Q} = \{q_g(\theta): g \in G\}$ and the Lie-group BLR reduces to (derivation in \Cref{app:multiplicative})
\begin{equation}\label{eq:multiplicativeUpdate}
	g_i \leftarrow g_i \exp \left( - \alpha \left(\E_{q_g}\left[\theta_i \partial_i \ell\right] - \tau\right) \right),
\end{equation}
where $\partial_i$ denotes the derivative with respect to $\theta_i$.
The derivation uses the facts that the differential in \Cref{eq:circled1formula} depends on $\E_{q_g}\left[\theta_i \partial_i \ell\right]$, the gradient of the entropy is simply 1, and the Fisher is again an identity matrix multiplied by a constant.
Notice that the parameter conditions $g_i>0$ are automatically satisfied, thus we stay on the manifold. The new algorithm can be used to train networks with desirable biologically-plausible attributes with sparse features. The new learning rule can be useful to design algorithms that encourage
explainability, compositionality, and disentanglement~\citep{BeZh20, WhDo22}.

We can show that linearization of \eqref{eq:multiplicativeUpdate} recovers the BLR-update for Rayleigh distributions \eqref{eq:rayleigh}. We show this for the 1-dimensional case. This is also an exponential family $\Qcal =\{q^{\lambda}(\theta) := \theta \lambda e^{-\frac{1}{2} \theta^2 \lambda} : \lambda >0\}$ where connection to \eqref{eq:rayleigh} is given as $\lambda = 1/g^2$. The Lie-group BLR can be written in terms of $\lambda$ by simply squaring and
reciprocating \eqref{eq:multiplicativeUpdate},
\begin{equation}
\begin{aligned}
	\lambda \leftarrow& \lambda \exp(2\alpha (\E_{q^\lambda}[\theta \, \partial \ell] - \tau)) \\
	&\approx \lambda + 2 \alpha \lambda(\E_{q^\lambda}[\theta \, \partial \ell ] - \tau),
      \end{aligned}
      \label{eq:recovertheblr}
    \end{equation}
    where the second line is using the linear approximation $e^x \approx 1+x$. We can show that this coincides with the BLR~\eqref{eq:rayblr} with a different step-size. 

To show this, we will simplify the BLR-update~\eqref{eq:BLR} for Rayleigh distributions. First, we can write the gradient $\nabla_\mu = F_\lambda^{-1} \nabla_\lambda$ where $F_\lambda$ is the Fisher. For the Rayleigh distributions, $F_\lambda = 1/\lambda^2$ and the entropy $\mathcal{H}(q^\lambda) = -\log \sqrt{\lambda} + \text{const}.$ We can write $\nabla_\lambda \E_{q^\lambda}[\ell]$ in terms of $\nabla_\theta \ell$ by using a change of variables $\theta \sqrt{\lambda} \mapsto \theta$ before differentiating.
Using these, we can simplify the BLR to get,
\begin{align}
   \lambda \leftarrow & \lambda + \frac{\alpha' \lambda}{2} \left(  \E_{q^\lambda} [\theta \, \partial \ell] - \tau \right) , \label{eq:rayblr}
\end{align}
which is same as \Cref{eq:recovertheblr} when step-size $\alpha = \alpha'/4$.

\begin{figure*}[t!]
  \centering
  \subcaptionbox{CIFAR-10 (CNN), additive, accuracy: $88.50$}{\includegraphics[width=0.49\linewidth]{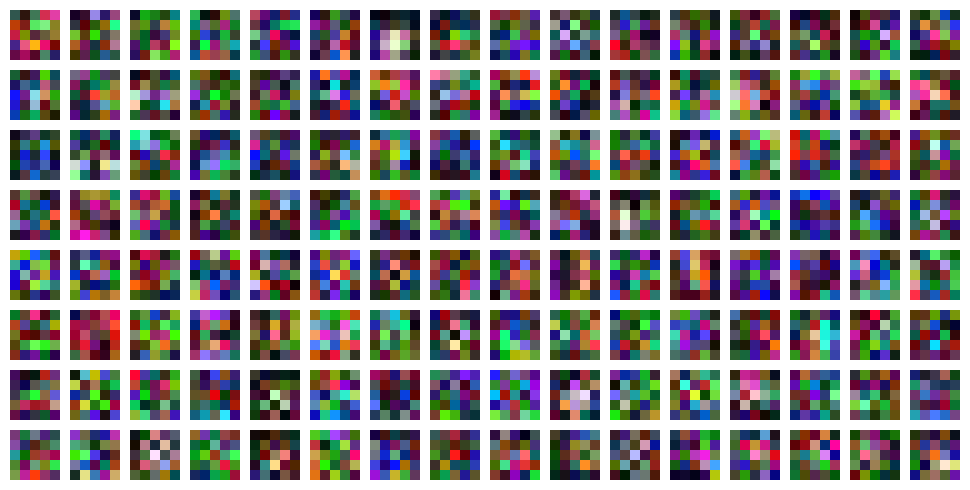}}
  \subcaptionbox{CIFAR-10 (CNN), multiplicative, accuracy: $87.55$}{\includegraphics[width=0.49\linewidth]{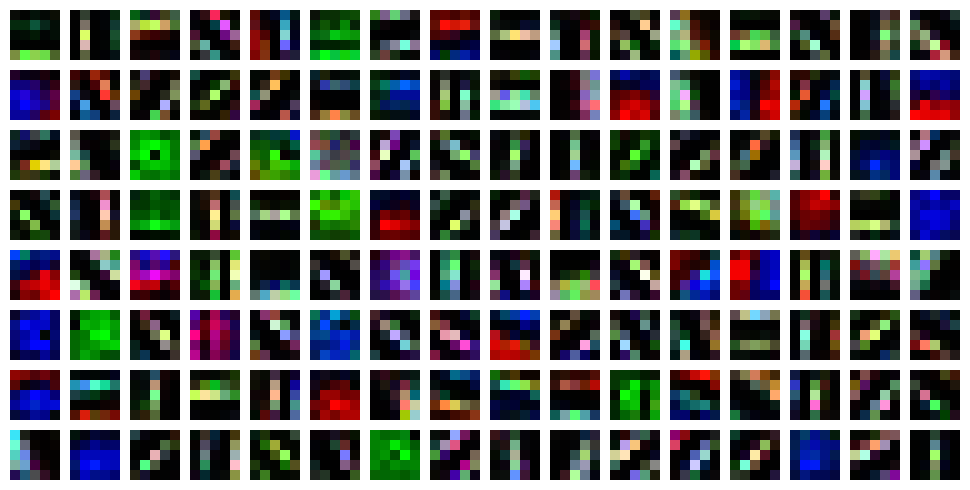}}
  \caption{Similar to the fully-connected setting of~\Cref{fig:filters_add_and_mult}, our multiplicative rule leads to sparse and interpretable weights (here: convolutional filters). The weights (filters) learned by the multiplicative rule have a shorter description length. For example, they can often be understood as edge detectors or are monochromatic (e.g., detecting blue or green patterns).}
  \label{fig:cnn}
\end{figure*}

\subsection{The diagonal affine group}
\label{sec:affine}

The action of the affine group combines translations and scaling. This group can be realized as pairs $(A, b)$ for diagonal positive $P\times P$ matrices $A$ and $b \in \R^P$ where the group operation is given by $(A_1,b_1) (A_2, b_2) = (A_1A_2, A_1b_2 + b_1)$ and the group action on $\Theta$ is given as $(A, b) \cdot \theta = A\theta + b$. 

The exponential map for this group is more complicated. Assuming ${q}_0$ is even (a technical assumption made only for a cleaner formula),  the update rule is (see \Cref{app:affine})
\begin{align}
	b_i &\leftarrow b_i + \frac{c_X}{c_y}A_i \frac{\exp\left(-\alpha U\right)-1}{U} V, \label{eq:diagAffineBUpdate}\\
	A_i &\leftarrow A_i \exp\left(-\alpha U\right), \label{eq:diagAffineAUpdate}
\end{align}
where $U = \E_{q_{g}}[(\theta_i - b_i) \partial_i \ell] - \tau$ and $V =A_i\E_{q_g}[\partial_i \ell] $. Also $c_X = \int_{\R} \left(1 + \theta\frac{\tilde{q}_0'(\theta)}{\tilde{q}_0(\theta)}\right)^2 \tilde{q}_0(\theta) \d \theta$, $c_y= \int_{\R} \frac{\tilde{q}_0'(\theta)^2}{\tilde{q}_0(\theta)}\d \theta$  are constants that can be calculated once and for all for a given $\tilde{q}_0$. Choosing $q_0$ as a Dirac delta measure gives us gradient descent as in the additive case.

In this group action if $q_0$ is chosen as the normal distribution $\mathcal{N}(0, I)$ then $\Qcal$ is also an exponential family. We show in \ref{app:KhanRue} that the linear approximation in $\alpha$ to this update is exactly the BLR from \citet{KhRu21}.

\begin{table}[t!]
  \resizebox{\linewidth}{!}{
	\begin{tabular}{c l r r r}
		\toprule
		\begin{tabular}{c}
		  \textbf{Model \&} \\
                  \textbf{Dataset}  \\
		\end{tabular} &
		\textbf{Method} & 
		\begin{tabular}{c}
                  \textbf{Accuracy} \textuparrow \\
                 {\scriptsize (higher is better)}
		\end{tabular} &
		\begin{tabular}{c}
			\textbf{NLL} \textdownarrow \\
                 {\scriptsize (lower is better)}
		\end{tabular} &
		\begin{tabular}{c}
                \textbf{ECE} \textdownarrow \\
                 {\scriptsize (lower is better)}
		\end{tabular} \\
                \midrule
		\multirow{2}{*}{
		  \begin{tabular}{c}
                    MNIST  \\
                    MLP
		\end{tabular}}   
		& add.~(\Cref{fig:additiveAlg})                           & $98.38_{\pm 0.02}$ & $0.083_{\pm 0.001}$ & $0.012_{\pm 0.000}$   \\
                & mult.~(\Cref{fig:multAlg})         & ${{98.59}}_{\pm 0.02}$ & ${{0.058}}_{\pm 0.001}$ & ${{0.006}}_{\pm 0.000}$    \\
                \midrule
		\multirow{2}{*}{
		  \begin{tabular}{c}
                    CIFAR--10  \\
                    MLP
		\end{tabular}}   
		& add.~(\Cref{fig:additiveAlg})                           &  $58.85_{\pm 0.08}$ & $1.236_{\pm 0.002}$ & $0.085_{\pm 0.001}$  \\
                & mult.~(\Cref{fig:multAlg})                                    &  ${{59.19}}_{\pm 0.07}$ & ${ 1.160}_{\pm 0.001}$ & ${ 0.026}_{\pm 0.001}$    \\
                \midrule
		\multirow{2}{*}{
		  \begin{tabular}{c}
                    CIFAR--10  \\
                    CNN
		\end{tabular}}   
		& add.~(\Cref{fig:additiveAlg})                       &  ${{88.50}}_{\pm 0.08}$ & $1.091_{\pm 0.007}$ & $0.096_{\pm 0.001}$  \\
                & mult.~(\Cref{fig:multAlg})                                    & $87.55_{\pm 0.06}$ & ${ 0.498}_{\pm 0.003}$ & ${ 0.034}_{\pm 0.001}$ \\
          \bottomrule \\
	\end{tabular}
  }
  \caption{Additive and multiplicative learning updates give comparable test accuracies.
  However, they learn very different representations, as shown in~\Cref{fig:filters_add_and_mult}. Moreover, learning with the multiplicative rule leads to a much smaller expected calibration error and negative log-likelihoods.}
\label{tab:mlpresults}
\end{table}

\section{NUMERICAL EXPERIMENTS}
\label{sec:numerics}

In this section, we compare our Lie-group BLR~\eqref{eq:updateRuleFleshedOut} to existing methods. 
We always report the performance for the predictive marginal probability $p(y \,|\, x)$. This can be computed from our optimal group element $g_* \in G$ via the
equation $p(y \,|\, x) = \int p(y \,|\, g_* \cdot \theta, x) \, q_0(\theta) \, \mathrm{d} \theta$. In practice,
we approximate the integral using $32$ samples independently drawn from $q_0$.

\subsection{Additive vs. multiplicative learning}
We now compare the properties of the additive and multiplicative group updates from \Cref{sec:additive} and \Cref{sec:multiplicative}
when applied to neural network training.
For a detailed pseudo-code of the final algorithm please see \Cref{fig:additiveAlg} and \Cref{fig:multAlg} in the appendix.
We use the additive and multiplicative updates to train a feed-forward neural network with 5 hidden layers (MLP) and a small convolutional net (CNN). The exact architectures and hyperparameters are described in \Cref{app:experiment1}. The results are summarized in~\Cref{tab:mlpresults}.
      \begin{table*}[t!]
	\centering
	\footnotesize
	\setlength{\tabcolsep}{4pt}
        \resizebox{\textwidth}{!}{
	\begin{tabular}{l  l | l l l | l l l | l  l  l }
		\toprule
		 \multirow{3}{*}{~~\textbf{Method}} &
		 \multirow{3}{*}{\textbf{Family $\mathcal{Q}$}}  &
		 \multicolumn{3}{|c|}{\textbf{CIFAR-10}} &  
		 \multicolumn{3}{|c|}{\textbf{CIFAR-100}} &  
		 \multicolumn{3}{|c}{\textbf{TinyImageNet}} \\
                                                  & &
                                                      		\begin{tabular}{c}
                  \textbf{Acc.} \textuparrow \\
                 {\scriptsize (higher is better)}
		\end{tabular} &
		\begin{tabular}{c}
			\textbf{NLL} \textdownarrow \\
                 {\scriptsize (lower is better)}
		\end{tabular} &
		\begin{tabular}{c}
                \textbf{ECE} \textdownarrow \\
                 {\scriptsize (lower is better)}
		\end{tabular} &
                                                                                      		\begin{tabular}{c}
                  \textbf{Acc.} \textuparrow \\
                 {\scriptsize (higher is better)}
		\end{tabular} &
		\begin{tabular}{c}
			\textbf{NLL} \textdownarrow \\
                 {\scriptsize (lower is better)}
		\end{tabular} &
		\begin{tabular}{c}
                \textbf{ECE} \textdownarrow \\
                 {\scriptsize (lower is better)}
		\end{tabular} &
                                \begin{tabular}{c}
                  \textbf{Acc.} \textuparrow \\
                 {\scriptsize (higher is better)}
		\end{tabular} &
		\begin{tabular}{c}
			\textbf{NLL} \textdownarrow \\
                 {\scriptsize (lower is better)}
		\end{tabular} &
		\begin{tabular}{c}
                \textbf{ECE} \textdownarrow \\
                 {\scriptsize (lower is better)}
		\end{tabular} 
		  \\ 
		 \midrule
		 \multirow{3}{*}{\begin{tabular}{c}Additive \\ (\Cref{fig:additiveAlg}) \end{tabular} }&
	         Uniform & $91.07_{\pm 0.08}$ & $0.365_{\pm 0.003}$ & $0.052_{\pm 0.001}$ & $64.29_{\pm 0.09}$ & $1.437_{\pm 0.004}$ & $0.121_{\pm 0.001}$    & $49.34_{\pm 0.14}$ & $2.234_{\pm 0.008}$ & $0.107_{\pm 0.001}$ \\
	         & Gaussian & $91.28_{\pm 0.11}$ & $0.328_{\pm 0.008}$ & $0.045_{\pm 0.001}$ & $64.61_{\pm 0.20}$ & $1.390_{\pm 0.008}$ & $0.107_{\pm 0.001}$    & $49.62_{\pm 0.15}$ & $2.204_{\pm 0.003}$ & $0.099_{\pm 0.002}$ \\
	         & Laplace & $91.14_{\pm 0.12}$ & $0.312_{\pm 0.005}$ & $0.039_{\pm 0.001}$ & $64.85_{\pm 0.13}$ & $1.359_{\pm 0.006}$ & $0.096_{\pm 0.001}$   & $49.73_{\pm 0.18}$ & $2.184_{\pm 0.007}$ & $0.089_{\pm 0.001}$ \\
		 \midrule
		 \multirow{3}{*}{\begin{tabular}{c}Affine \\ (\Cref{fig:affineAlg}) \end{tabular}}&
	         Uniform & $91.60_{\pm 0.05}$ & $0.300_{\pm 0.002}$ &  $0.040_{\pm 0.001}$ & $66.08_{\pm 0.12}$ & $1.288_{\pm 0.007}$ & $0.093_{\pm 0.002}$    & $51.19_{\pm 0.12}$ & $2.099_{\pm 0.005}$ & $0.076_{\pm 0.001}$ \\
	         & Gaussian & $91.53_{\pm 0.10}$ & $0.294_{\pm 0.004}$ & $0.036_{\pm 0.001}$ & $66.55_{\pm 0.10}$ & $1.255_{\pm 0.005}$ & $0.079_{\pm 0.002}$    & $51.13_{\pm 0.16}$ & $2.098_{\pm 0.004}$ & $0.070_{\pm 0.002}$ \\
                                                    & Laplace & ${91.87}_{\pm 0.04}$ & ${0.272}_{\pm 0.002}$ & $0.029_{\pm 0.001}$  & $66.44_{\pm 0.10}$ & $1.247_{\pm 0.006}$ & $0.071_{\pm 0.001}$    & $51.36_{\pm 0.14}$ & $2.101_{\pm 0.009}$   & $0.065_{\pm 0.001}$ \\
		 \midrule
	         ~~~SGD & -- & $91.22_{\pm 0.07}$ & $0.354_{\pm 0.006}$ & $0.050_{\pm 0.001}$ & $64.19_{\pm 0.14}$ & $1.431_{\pm 0.007}$  & $0.121_{\pm 0.001}$    & $49.48_{\pm 0.10}$ & $2.231_{\pm 0.004}$ & $0.106_{\pm 0.001}$ \\
	         ~~~iVON & Gaussian & ${91.80}_{\pm 0.05}$ & $0.288_{\pm 0.003}$ & $0.038_{\pm 0.001}$ & $66.59_{\pm 0.11}$ & $1.209_{\pm 0.004}$ & $0.049_{\pm 0.001}$    & ${52.13}_{\pm 0.08}$ & ${1.982}_{\pm 0.003}$ & ${0.018}_{\pm 0.001}$ \\
	         ~~~VOGN & Gaussian & $91.32_{\pm 0.09}$ & ${0.264}_{\pm 0.003}$ & ${0.011}_{\pm 0.000}$ &  ${{66.81}}_{\pm 0.14}$ & ${1.183}_{\pm 0.007}$ & ${0.020}_{\pm 0.002}$    & $51.09_{\pm 0.12}$ & $2.045_{\pm 0.006}$ & ${0.016}_{\pm 0.001}$ \\
		\bottomrule
	\end{tabular} 
      } 
        \caption{Our proposed affine learning rule~(\Cref{fig:affineAlg}) performs competitively to SGD and state-of-the art Bayesian approaches \textbf{VOGN}~\citep{OsSw19} and the Adam-like optimizer from \citet[Figure~1]{LiSc20} which we refer to as \textbf{iVON}. Empirically, making the $q_0$ distribution more heavy tailed (going from uniform to Gaussian, and from Gaussian to Laplace) improves calibration measures like ECE and NLL.}
	\label{table:additive_vs_affine}
\end{table*}

While both the additive (Gaussian $q_0$) and multiplicative updates (Rayleigh $q_0$) lead to comparable
test accuracies on the MNIST and CIFAR-10 data sets, the learned neural network weights are drastically different. Multiplicative learning leads
to sparse, localized and compositional traits. This highlights how different
choices of Lie group can lead to different learning behaviors. 
The weights of the trained neural networks are visualized in \Cref{fig:filters_add_and_mult} for the MLP and in \Cref{fig:cnn} for the CNN.
Multiplicative learning also tends to improve the negative-log likelihood (NLL) as well as the expected calibration error (ECE)~\citep{GuPl17}.

The sparse nature of the filters in the multiplicative family can be explained as the effect of entropy and the mean of a distribution being intimately tied together. Weight distributions of connections with large mean also are spread out, i.e., have large entropy. Therefore in avoiding large expected errors, any unnecessary non-robust weight magnitude is suppressed. We may interpret the resulting sharpness of the filters as neuronal task specialization in attributes we humans can convey such as color, location and orientation. For example, the readers probably can locate the multiplicative filters in~\Cref{fig:filters_add_and_mult} when referred to simply as ``\emph{the blue dot in the bottom right}'' or ``\emph{multicolor vertical stripe left of center}''. For the representations learned by the additive rule, no such short descriptions exist.

\subsection{The affine learning rule}
Finally, we compare our affine learning update from~\Cref{sec:affine} to state-of-the-art natural-gradient variational inference methods: VOGN~\citep{OsSw19} and the Adam-like optimizer given in~\citet[Figure~1]{LiSc20} which we refer to as iVON. For a detailed pseudo-code of our affine update rule, see~\Cref{fig:affineAlg} in the appendix. The comparison is carried out for a standard ResNet-20 architecture which reaches around $91 \%$ when trained with SGD, see~\citep[Table~6]{HeZh16}. The hyperparameters and other details are in~\Cref{app:experiment2}.

\Cref{table:additive_vs_affine} summarizes our results. Our algorithm yields competitive results to SGD, VOGN and iVON, yet offers more flexibility in the choice of distribution: VOGN and iVON are updating a Gaussian distribution, whereas our method works for any base distribution $q_0$. Using a heavy-tailed Laplace distribution leads to improvements in NLL and ECE compared to a Gaussian or the even more thin-tailed uniform distribution. Moreover, both VOGN and iVON require a small additional
damping term to stabilize the learning algorithm, see~\citet{OsSw19}. Our algorithm does not require any such additional term, and is easier to tune.

The learning update rule arising from an additive group (\Cref{sec:additive}) has been recently studied by~\citet{OrKe22} in the context of regularizing noise injections. Using an affine update allows one to learn the variance of the noise. \Cref{table:additive_vs_affine} shows that this leads to improvements while eliminating an additional hyperparameter which controls the strength of the noise.
Additive and affine columns were computed with 1 MC sample only, thus the compute cost is comparable to that of SGD.

\section{DISCUSSION}
\label{sec:discuss}
We propose the Lie-group BLR which extends the BLR by using Lie-groups and can be much easier to use in many cases. Unlike the BLR, the new rule does not rely on a specific parameterization of EFs, enables
gradient computations via a general yet easy-to-use reparametrization trick, and automatically keeps the updates on the manifold. We show three use cases of the new rule for algorithm design in deep learning, including a new algorithm for training networks with biologically-plausible attributes.

Our work clearly shows the usefulness of Lie-groups but more work is needed in identifying and characterizing the class of distributions where the new rule is easy to use. For example, we have shown 3 cases where the BLR is a coarse linear-approximation of the new rule, but is this true for all minimal EFs? Similarly, for what distributions does the Fisher computation remain easy?
Same question for the exponential map.
More work is needed to answer these questions. Another interesting direction is to use the new rule for the design of better algorithms in deep learning, for example, those focusing on explainability, compositionality, and disentanglement.

\subsubsection*{Acknowledgements}
We would like to thank Koichi Tojo (RIKEN AIP), Akiyoshi Sannai (RIKEN AIP), Asuka Takatsu (Tokyo Metropolitan University), Beno\^it Collins (Kyoto University) and Kenichi Bannai (RIKEN AIP \& Keio University) for various discussions and feedback.
This work was 
supported by the Bayes-duality project, JST CREST Grant Number JPMJCR2112. Eren Mehmet K{\i}ral was supported by the RIKEN Special Postdoctoral Researcher Program.

\subsubsection*{Author Contributions Statement}
List of Authors:  Eren Mehmet K{\i}ral (EMK), Thomas M\"ollenhoff (TM), Mohammad Emtiyaz Khan (MEK).

Based on in-depth discussions with TM, EMK proposed the Lie group framework, derived the specific algorithms and the connections to existing methods. MEK provided feedback on these. TM designed and conducted the experiments with suggestions from EMK and MEK. MEK and EMK wrote the paper together, with feedback from TM.

\bibliography{./refthomas}


\clearpage
\appendix

\thispagestyle{empty}

\onecolumn

\section{MATHEMATICAL DETAILS}

\subsection{The differentials and the Fisher metric}
\label{app:Fisher}

\subsubsection{Fisher is the second differential of KL-Divergence}

The Fisher metric can be obtained as the second order differential of our objective function $\mathcal{E}(q) = \DKL(q\|p_\tau)$. Let $h \in T_q\Qcal$ be a tangent vector and let us perturb the energy functional by a mean zero function $h$. Its second order approximation is given by
\begin{equation}\label{eq:EquadraticExpansion}
	\mathcal{E}(q + h) \approx \mathcal{E}(q) + (\mathrm{d} \mathcal{E}(q))[h] + \frac12 (\mathrm{d}^2 \mathcal{E}(q))[h,h].
\end{equation}
Let us give a sketch of calculation the terms in this quadratic expansion of $\mathcal{E}$, only giving the main idea, as the derivation of the second differential of KL-divergence is already well known, see, e.g. John Baez's blog \url{https://math.ucr.edu/home/baez/information/information_geometry_7.html} (as of Oct 11, 2022). Perturb by $h$,
\begin{align*}
	\mathcal{E}(q + h) &= \DKL(q + h\|p_\tau) = \int (q + h) \log\left(\frac{q + h}{p_\tau}\right) \\
	&= \int (q + h) \log \left(\frac{q}{p_\tau}\right) + \int (q + h) \log \left(\frac{q + h}{q}\right) \\
	&= \underbrace{\int q \log\left(\frac{q}{p_\tau}\right)}_{\mathcal{E}(q)} + \int h \log \left(\frac{q}{p_\tau}\right) + \int (q + h) \left(\frac{h}{q} - \frac{h^2}{q^2} + \cdots \right), 
\end{align*}
where in the second line we multiplied and divided the ratio inside the logarithm by $q$ and in the last line we applied the Taylor expansion of $\log (1 + x)$ with $x = h/q$. Continuing the calculation we get that
\[
	\mathcal{E}(q + h) - \mathcal{E}(q) =  \int h \log \left(\frac{q}{p_\tau}\right) + \cancelto{0}{\int h }+ \frac{1}{2}\int \frac{h^2}{q} + \cdots
\]
The linear term must be the first differential and the quadratic term is the second differential $(\mathrm{d}^2\mathcal{E}(q))[h,h]$. Using polarization identities we can get the quadratic term as a bilinear form
\begin{equation}\label{eq:appEFirstAndSecondDifferentials}
	(\mathrm{d}\mathcal{E}(q)) [h] = \int_{\Theta} h(\theta) \log \left( \frac{q(\theta)}{p_\tau(\theta)} \right) \d \theta \qquad \text{ and } \qquad (\mathrm{d}^2\mathcal{E})(q))[h_1,h_2] = \int_{\Theta} \frac{h_1(\theta)}{q(\theta)}\frac{h_2(\theta)}{q(\theta)} q(\theta) \d \theta.
\end{equation}
The second differential can also be written in the form $\E_q[(\partial_{h_1} \log q) (\partial_{h_2}\log q)]$, this is exactly the Fisher metric. Therefore one can see that choosing the Fisher metric as the direction of fastest descent is also compatible with minimizing the quadratic expansion of the objective function $\mathcal{E}$. 

\subsubsection{Fisher metric is independent of base point}

If the tangent vectors are parametrized by the Lie algebra as $h_X$ then we can write them as $\pi(g)$ of a tangent vector at identity: $$h_X = \frac{\d}{\d t} \pi (ge^{tX}) q_0 \bigg|_{t = 0} = \pi(g) \frac{\d}{\d t} q_{e^{tX}}\bigg|_{t = 0} = \pi(g) h_X^e.$$ This means that by making a change of variables $\theta \mapsto g\cdot \theta$ the we get a quantity that is independent of $g$. Indeed
\begin{align*}
	\omega_{\text{Fisher}}(h_{X}^{g}, h_{Y}^{g} )  &= \int_{\Theta} \frac{h_{X}^{g}(\theta)}{q_g(\theta)} \frac{h_{Y}^{g}(\theta)}{q_g(\theta)} q_g(\theta) \d \theta = \int_{\Theta} \frac{h_{X}^{e}(g\inv\cdot  \theta)}{q_0(g\inv \cdot \theta)} \frac{h_{Y}^{e}(g\inv \cdot \theta)}{q_0(g\inv \cdot \theta)} \left|\frac{\d(g\cdot \theta)}{d\theta}\right|\inv q_0(g\inv\cdot  \theta) \d \theta \\
	&= \int_{\Theta} \frac{h_{X}^{e}(\theta)}{q_0(\theta)} \frac{h_{Y}^{e}(\theta)}{q_0(\theta)}  q_0(\theta) \d \theta  = \omega_{\text{Fisher}}(h_X^e, h_Y^e).
\end{align*}
for any $X,Y \in T_eG$ and $g \in G$. In the second line we made use of a change of variables $\theta \mapsto g\cdot \theta$. This is simply a bilinear form in the Lie algebra $T_eG$. So, it is enough to compute the Fisher metric once and for all, and we do not need to compute a different metric at each point throughout the training.

\subsection{Differential of the entropy term in $\mathrm{d}\mathcal{E}(q)$}
\label{app:DEcalculation}

The entropic contribution $\circled{2}$ to the differential can be calculated in a similar fashion to $\circled{1}$, as shown below,
\begin{align}
	\circled{2} &= \frac{\d}{\d t} \left(\int q_{g e^{tX}} (\theta) \log q_{g} (\theta) \d \theta \right)\Bigg|_{t = 0} \notag\\
	&= \frac{\d}{\d t}\left( \int q_0(\theta) \log \left(\frac{1}{|\d(g\cdot \theta)/\d\theta|} q_0\left( e^{tX} \cdot\theta  \right) \right) \d \theta \right)\Bigg|_{t = 0} \notag \\
	&= \frac{\d}{\d t} \bigg(\cancelto{0}{-\int \log(|\d(g\cdot \theta)/\d \theta|) q_0(\theta)} + \int q_0(\theta) \log q_0\left(e^{tX} \cdot \theta \right) \d \theta  \bigg)\Bigg|_{t=0} \notag \\
	& = \int  \left(\nabla_{\theta} q_0(\theta)\right)^\top \frac{\d}{\d t} \left(e^{tX}\cdot \theta \right)\bigg|_{t = 0} \d \theta = \int  \left(\nabla_{\theta} q_0(\theta)\right)^\top X\cdot \theta \d \theta.\label{eq:circled2calculations}
\end{align}
On the first line we use the definition of the tangent vector $h_X$, the second line makes a change of variables $\theta \mapsto ge^{tX} \cdot \theta$. We can cancel the $|\d (g\cdot \theta)/\d \theta|$ term in the third line because it has no $t$ dependence, and on the last line we go one step further and apply the chain rule, cancelling the $q_0$ term via the logarithmic derivative. The definition of the action of $T_eG$ on $\Theta$ is given exactly as the linearization i.e. $X \cdot \theta := \frac{\d}{\d t} e^{tX} \cdot \theta \big|_{t =0}$.

\subsection{Specializing the  update rule to the additive group}
\label{app:additive}

\begin{algorithm}[h!]
   \caption{Additive Update}
      \label{fig:additiveAlg}
      \definecolor{commentgray}{rgb}{0.25, 0.25, 0.25}
      \begin{algorithmic}[1]
          \REQUIRE $\alpha>0$ step size, $K\geq 1$ MC-sample number, $\beta \in [0, 1)$, $\nu$ a distribution on $\R$.
          \STATE Initialize $g \in \R^P$ randomly, and $M  = \mathbf{0} \in \R^P$.
          \WHILE{not converged}
          \STATE Sample a minibatch $\mathcal{J} \subseteq [1..N]$ of size $n$,
          \STATE Sample noise vectors $\varepsilon_k \sim \nu^P$ for $k = 1, \ldots, K$,
          \STATE Put $\theta_k = g + \varepsilon_k$, the MC-parameter samples,
          \STATE $ U \leftarrow \frac{1}{K} \sum_{k=1}^K \left(\frac{1}{n} \sum_{j \in \mathcal{J}} \nabla_\theta \ell_j(\theta_k) + \frac{1}{N} \nabla_\theta R(\theta_k)\right)$,
          \STATE $M \leftarrow (1 - \beta) U + \beta M$,           \STATE $g \leftarrow g - \alpha M$,
          \ENDWHILE
          \STATE Return $g$.
        \end{algorithmic}
        \emph{Loss function: $\ell(\theta) = \sum_{i=1}^N \ell_{i}(\theta) + R(\theta)$.}
\end{algorithm}

In this case $\Theta = \R^P$ with $G = (\R^P, +)$ acting by $g \cdot \theta = \theta + g$ on the space of parameters $\Theta$. The tangent bundle of $G$ is trivial, with each tangent space isomorphic to $\R^P$. The exponential map for $X \in T_g(\R^P) \cong \R^P$ is simply given by identity, i.e.\ $\exp (X) = X$.

We work in the mean-field case. That is, let $\tilde{q}_0$ be the density function of an everywhere positive probability distribution on $\R$, and put $q_0(\theta) = \prod_i \tilde{q}_0(\theta_i)$. By making this choice, we assume that the probability for each parameter is independent and identically distributed. Taking the base point distribution $q_0$ the orbit of $q_0$ under under the action of $G$ via pushforwards gives us
$
	\Qcal := \{q_g(\theta)\d \theta : g \in \R^P \}
$ where $q_g(\theta) := (\pi(g)\cdot q_0)(\theta) = q_0(\theta -g)$.

The infinitesimal action of $T_eG$ on $\Theta$ is given as $X \cdot \theta = \frac{\d}{\d t} (\exp(tX) \cdot \theta)\big|_{t = 0} = \frac{\d}{\d t} (\theta + tX) \big|_{t = 0} = X \in T_\theta \Theta$. And since the group is commutative, the adjoint representation is trivial, i.e.\ $\operatorname{Ad}_g(X) \cdot \theta = X\cdot \theta = X$. 

The differential of $E$ can be calculated by \eqref{eq:DEsplit}. First note that the integral \circled{2} in \eqref{eq:circled2formula} vanishes as we get $\circled{2} = \left(\int \nabla_{\theta}q_0(\theta) \d \theta \right)^\top X = \mathbf{0}^\top X = 0$. Indeed upon integration by parts
\[
	\int_{\Theta} \partial_i q_0(\theta) \d \theta = \int_{\R} \tilde{q}_0'(\theta_i) \d \theta_i \times \prod_{j \neq i} \int_{\R} \tilde{q}_0(\theta_j) \d \theta_j = \tilde{q}_0(\theta_i)\bigg|_{\theta_i = -\infty}^\infty = 0-0 = 0.
\] This fact should not be surprising since the group action only translates the mean of the distribution $q_0$ and there is no change in entropy. As for \eqref{eq:circled1formula} we may again take the $X$ dependence out and write the integral as  an expectation. Thus the differential is calculated as
\[
	(\mathrm{d}\mathcal{E}(q_g))[h_X] = \circled{1} = \E_{q_g}[\nabla_\theta \ell]^\top X.
\]
Since $\operatorname{Ad}_g(X) \cdot \theta = X \cdot \theta = X$. Now we calculate the Fisher information matrix, to so that we may apply the musical-isomorphism $\sharp$ and get an element of $T_{q_g}\Qcal$ instead of the covector $\mathrm{d}\mathcal{E}(q_g)\in T_{q_g}^*\Qcal$. 

The tangent vectors to $\Qcal$ are given as mean-zero functions. More concretely in our case  at any $q = q_g$ the tangent space is spanned by the 
\[
	h_i(\theta) = (\partial_i q_0)(\theta - g)
\]
and with respect to this basis the Fisher information metric is calculated as,
\[
	\omega_{\text{Fisher}}(h_i, h_j) = \int_{\R^P} \frac{\partial_i q_0(\theta)}{q_0(\theta)} \frac{\partial_j q_0(\theta)}{q_0(\theta)} q_0(\theta) \d \theta = \int_{\R^P} \frac{\tilde{q}'_0(\theta_i)}{\tilde{q}_0(\theta_i)} \frac{\tilde{q}'_0(\theta_j)}{\tilde{q}_0(\theta_j)} q_0(\theta) \d \theta.
\]
For $i \neq  j$ this reduces to $\left(\int_{\R} \tilde{q}_0'(\theta)\d \theta\right)^2 = 0$, and so $F$ is given by a scalar matrix $F = c_F I_{P\times P} $ where the constant $c_F > 0$ is given by the integral
\[
	c_F = \int_\R \frac{\tilde{q}_0^\prime (\theta)^2}{\tilde{q}_0(\theta)} \d \theta.
\]

The update rule in this case is then given by
\begin{equation} 
	g^{\text{updated}} = g + \exp\left(- \alpha \E_{q_g}[\nabla_\theta \ell]\right) =  g - \alpha \E_{q_{g}}[\nabla_\theta \ell]
\end{equation}
for some step size $\alpha > 0$, where we have absorbed the Fisher constant $c_F$ into the step size. Note that absorbing the Fisher constant into the step-size is exactly why we are able to then substitute distributions such as Dirac delta as $q_0$.

\subsection{Specializing to the multiplicative group}
\label{app:multiplicative}

\begin{algorithm}[h!]
   \caption{Multiplicative Update}
      \label{fig:multAlg}
      \definecolor{commentgray}{rgb}{0.25, 0.25, 0.25}
      \begin{algorithmic}[1]
          \REQUIRE $\alpha>0$ step size, $K\geq 1$ MC-sample number, $T$ temperature, $\beta \in [0, 1)$, $\nu$ a distribution on $\R_{>0}$.
          \STATE Initialize $g \in \R_{>0}^P$ randomly, and $M = \mathbf{0} \in \R^P$.
          \WHILE{not converged}
          \STATE Sample a minibatch $\mathcal{J} \subseteq [1..N]$ of size $n$,
          \STATE Sample noise vectors $\varepsilon_k \sim \nu^P$ for $k = 1, \ldots, K$,
          \STATE Put $\theta_k = g \varepsilon_k$, the MC-parameter samples,
          \STATE $U_k = \frac{1}{n} \sum_{j \in \mathcal{J}} \theta_k \nabla_\theta \ell_j(\theta_k) + \frac{1}{N} \theta_k \nabla_\theta R(\theta_k) - \frac{\tau}{N}\mathbf{1},$
          \STATE $U \leftarrow \frac{1}{K} \sum_{k = 1}^K U_k$
          \STATE $M \leftarrow (1 - \beta) U + \beta M$,           \STATE $g \leftarrow g \exp(- \alpha M)$
          \ENDWHILE
          \STATE Return $g$.
     \end{algorithmic}
	\emph{Loss function: $\ell(\theta) = \sum_{i=1}^N \ell_{i}(\theta) + R(\theta)$. Here, $ab$ for two vectors $a,b \in \R^P $ is taken to mean componentwise multiplication and $\exp$ is also computed componentwise. }
\end{algorithm}

For the multiplicative case we have $\Theta = \R_{>0}^P$. Fixing the signs of a neural network's weights we obtain such a model, where the (positive) magnitudes of the weights become the parameters of the model. The group $G = \R_{>0}^P$, considered with componentwise multiplication as the group operation acts on $\Theta$ in the same way  $g \cdot \theta := g \theta$. Here the componentwise product of two vectors $a,b$ is simply denoted by $ab$.

Again at a mean-field base distribution $q_0(\theta) = \prod_{k = 1}^P \tilde{q}_0(\theta_k)$, the orbit of the pushforward measures look $\Qcal = \{q_g = \pi(g) \cdot q_0 : g \in G\} = \{\frac{1}{|g|} q_0(g\inv \theta) : g \in G\}.$
Here $|g| = \prod_{k = 1}^P g_i$ is the determinant of the Jacobian $|\d(g\cdot \theta)/\d\theta|$.

\subsubsection{Examples}
This scheme includes important families. 
\begin{itemize}
\item Choose $\widetilde{q}_0 (\theta) = e^{-\theta}$, then we get the family of exponential distributions
\[
	\pi(g) \cdot q_0(\theta) = \prod_k \frac{1}{g_k} e^{-\theta_k/g_k}
\]
is the family of exponential distributions $\{\prod_i \lambda_i e^{- \langle \theta, \lambda\rangle }: \lambda \in \R_{>0}^{P}\}$. The group parameter $g$ and the natural parameter $\lambda$ are componentwise reciprocals of each other.

\item
Choosing $\widetilde{q}_0(\theta) = \theta e^{-\frac12 \theta^2}$  gives us the family of Rayleigh distributions
\[
	\pi(g) \cdot q_0(\theta) = \prod_k \frac{\theta_i}{g_i^2} e^{-\theta_i^2/(2g_i^2)}.
\]
In this case the $\sigma_i$ parameter of the Rayleigh distribution exactly match up with $g_i$. 
\item Log-normal distributions with a fixed variance parameter also fit into this family scheme. Let us put 
\[
	\widetilde{q}_0 (\theta) = \frac{1}{\theta\sigma \sqrt{2\pi}} e^{-\frac{(\ln \theta - m)^2}{2\sigma^2}}
\]
Then if $g > 0$ we get
\[
	\frac{1}{g}\widetilde{q}_0(g\inv \theta) = \frac{1}{\theta \sigma \sqrt{2\pi}} e^{- \frac{(\ln \theta - \ln g-m)^2}{2\sigma^2}},
\]
the log-normal distribution with mean $m + \ln g$ and the same scale parameter $\sigma$. The action of $\pi(g)$ translates the parameter $m$ by $\ln g $.

\end{itemize}

\subsubsection{The tangent vectors}
The Lie algebra of $G$ is given by vectors $X \in \R^P$. Using that we can parametrize the tangent space of $\Qcal$ at any $q_g$. 

\begin{lemma}\label{lem:tangentBasis}
Given a $q = q_g \in \Qcal$ we have a basis of tangent vectors given as functions on $\theta$. They integrate to $0$, and are explicitly given by
\[
	h_i(\theta) =- \pi(g) \cdot \left(q_0(\theta)\left(1 + \theta_i \frac{\tilde{q}_0'(\theta_i)}{\tilde{q}_0(\theta_i)}\right)\right).
\]
\end{lemma}

\begin{proof}
We calculate from the definition.
\begin{align*}
h_X^g(\theta) &= \frac{\d}{\d t} q_{g \exp(tX))}(\theta) \bigg|_{t = 0} \\
&= \frac{d}{\d t} \left( \frac{1}{|g \exp(tX)|} q_0( \exp(-tX) g\inv  \cdot \theta) \right)\Bigg|_{t = 0}\\
	& =  -\frac{(\sum_i X_i)}{|g|} q_0(g\inv \theta) - \sum_i \frac{1}{|g|}\partial_i q_0(g\inv \theta) X_i \frac{\theta_i}{g_i}.
\end{align*}
Here $\exp$ means that we apply exponentiation componentwise, as well as product of two vectors. 
Substituting the standard basis $e_i$ for $X$ gives a basis for $T_{q_g}\Qcal$. Let us call it $h_1, h_2, \ldots, h_P$. The above formula can then be written more succinctly as
\[
	h_i(\theta) = - q_g(\theta) \left( 1 + \frac{\theta_i}{g_i} \frac{\partial_i q_0(g\inv  \theta)}{q_0(g\inv  \theta)} \right) = - q_g(\theta) \left(1 + \frac{\theta_i}{g_i} \frac{{\tilde{q}_0}^\prime(\frac{\theta_i}{g_i})}{\tilde{q}_0(\tfrac{\theta_i}{g_i})}\right).
\]
Noting the implicit $\pi(g)$ action, we get the result.
\end{proof}

Another way to write these tangent vectors are $h_i(\theta) = -q_g(\theta)- \pi(g) \delta_i q_0(\theta)$ where $\delta_i$ are the invariant differential operators on functions on $\Theta$ given by $\delta_i f(\theta) = (\theta_i \partial_i f)(\theta).$
The reason this is called an \emph{invariant} operator is because it is invariant under the group action by $G$, in other words $\delta_i \pi(g) f = \pi(g) \delta_i f$ for all $i =1, \ldots, P$.

\subsubsection{The Fisher metric}

 We now calculate the Fisher metric as a matrix with respect to the basis of tangent vectors given above in Lemma \ref{lem:tangentBasis}. 

\begin{lemma}\label{lem:multFisher}
	The matrix for the Fisher bilinear form with respect to the given basis $h_1, \ldots, h_P \in T_{q_g}\Qcal$ above, is $c_F I$. Here $I = I_{P\times P}$ is the identity matrix and $c_F>0$ is a constant that depends only on $\tilde{q}_0$. 
\end{lemma}

Notice that this metric does not depend on $g$, i.e.\ with this parametrization it is independent of the basepoint $q_g$. 

\begin{proof}
The Fisher bilinear form is $\int (\partial_{h_i} \log q_g(\theta) ) (\partial_{h_j} \log q_g(\theta)) q_g(\theta) \d \theta$ by definition. Here $\partial_h$ means that we are taking a directional derivative in the space of all measures in the direction of $h$. We calculate,
\begin{align*}
	\omega_{\text{Fisher}}(h_i, h_j) & = \int \frac{h_i(\theta)}{q_g(\theta)} \frac{h_j(\theta)}{q_g(\theta)} q_g(\theta) \d \theta\\
	&=  \int \left(1 + \frac{\theta_i}{g_i} \frac{{\tilde{q}_0}^\prime(\frac{\theta_i}{g_i})}{\tilde{q}_0(\tfrac{\theta_i}{g_i})}\right) \left(1 + \frac{\theta_j}{g_j} \frac{{\tilde{q}_0}^\prime(\frac{\theta_j}{g_j})}{\tilde{q}_0(\tfrac{\theta_j}{g_j})}\right) q_g(\theta) \d \theta\\
	&= \int \left(1 + \theta_i \frac{{\tilde{q}_0}^\prime(\theta_i)}{\tilde{q}_0(\theta_i)}\right) \left(1 + \theta_j \frac{{\tilde{q}_0}^\prime(\theta_j)}{\tilde{q}_0(\theta_j)}\right) q_0(\theta) \d \theta.
\end{align*}
In the last line we made a change of variables $\theta \mapsto g\cdot\theta$. 
Now there are two cases, firstly if $i \neq j$, 
\begin{align*}
	\omega_{\text{Fisher}}(h_i, h_j) &= \prod_{k \neq i, j} \int\limits_{0}^\infty \widetilde{q}_0(\theta_k) \d \theta_k \left(\int\limits_{0}^\infty (\tilde{q}_0(\theta_i) + \theta_i {\tilde{q}_0}^\prime(\theta_i)) \d \theta_i\right)
	\left(\int\limits_0^\infty (\widetilde{q}_0(\theta_j) + \theta_j {\tilde{q}_0}^\prime(\theta_j)) \d \theta_j \right) \\
	&= \left(\int_0^\infty \frac{\d}{\d\theta} \left(\theta \tilde{q}_0(\theta)\right) \d \theta \right)^2 = \left(\theta \tilde{q}_0(\theta) \Big|_{t = 0}^\infty \right)^2 = 0.
\end{align*}

Thus with respect to this basis, the matrix of the Fisher metric is diagonal. The value at these diagonal elements is calculated via
\[
	c_F := \int_{0}^\infty \left(1 + \theta \frac{\tilde{q}_0^\prime(\theta)}{\tilde{q}_0(\theta)}\right)^2 \tilde{q}_0(\theta) \d \theta.
\]
This is a nonnegative number. Most importantly it is independent of $g$ and only depends on the $\tilde{q}_0$ we chose.
\end{proof}

 The constant of this lemma is given as  $c_F = 1$ if $\widetilde{q}_0(\theta) = e^{-\theta}$ giving the exponential distributions, and $c_F = 4$ if $\widetilde{q}_0(\theta)= \theta e^{-\frac{\theta^2}{2}}$. In general $c_F$ is clearly nonnegative, but above we claimed more. That it was positive. The only way the integral could be zero is if $\tilde{q}_0$ satisfies the differential equation $\tilde{q}_0'(\theta) = -\tilde{q}_0(\theta)/\theta$ which has solutions $\tilde{q}_0 = c/\theta$. Notice that these solutions do not have finite integrals on $(0,\infty)$ and thus fall outside our purview.

\subsubsection{The Differential of $\mathcal{E}$}

We apply \eqref{eq:DEsplit} and the calculations below that to our specific situation.	

\begin{lemma}\label{lem:EDifferential}
Let $h_{X} = \sum_i X_i h_i$ be a tangent vector in $T_{q_g}\Qcal$.Then the differential of $\mathcal{E}$ at $q_g$ evaluated at $h_{X}$ is given as 
\[
	(\mathrm{d}\mathcal{E}(q_g))[h_X] = n^\top X \quad \text{ where } \quad n = \frac{1}{\tau}\E_{q_g}\left[\theta \nabla_\theta \ell\right] - \mathbf{1}.
\]
Here $\mathbf{1} = \begin{bmatrix} 1& 1& \cdots &1 \end{bmatrix}^\top$. 
\end{lemma}

\begin{proof}

Note that $T_eG \cong \R^P$ and exponential map is componentwise exponentiation: $e^{X} = (e^{X_1}, e^{X_2}, \ldots, e^{X_P}) \in G$. Therefore the infinitesimal action of the Lie algebra on the parameters is given as $X \cdot \theta = \tfrac{\d}{\d t} (e^{tX} \cdot \theta)\big|_{t = 0} = X \theta$ (again, understood as componentwise multiplication). Also as the group is abelian $\operatorname{Ad}_g(X) = X$. We have
\begin{equation*}
\circled{1} =  \int q_g(\theta) (\nabla_\theta \ell(\theta))^\top (X \theta)  \d \theta
= \sum_{i=1}^P X_i \int q_g(\theta) \theta_i \partial_i \ell(\theta) \d\theta =  \E_{q_g}[\theta \nabla_{\theta} \ell]^\top X
\end{equation*}

For the second part
\begin{equation*}
	\circled{2} = \sum_i X_i \int \theta_i(\partial_i q_0)(\theta) = \sum_{i} X_i \int_0^\infty \theta_i \tilde{q}_0'(\theta_i) \d \theta_i
\end{equation*}
where on the last equality we used the multiplicative structure of $q_0$. The resulting integral can be calculated via integration by parts as
\[
	\int_0^\infty \theta \tilde{q}_0' (\theta) \d \theta = \theta \tilde{q}_0(\theta)\Big|_{0}^\infty - \int_0^\infty \tilde{q}_0(\theta)\d \theta = -1.
\]
The contribution from this part will be $-\mathbf{1}^\top X$. Therefore $$(\mathrm{d}\mathcal{E}(q_g))[h_X] = \left(\frac{1}{\tau} \E_{q_g}[\theta \nabla_\theta \ell]^\top - \mathbf{1}\right)^\top X$$ as claimed.
\end{proof}

\subsubsection{The update rule}

We combine results of the previous section in order to write down the explicit update rule on $\Qcal$. 

Firstly let us note that we are looking for a vector $n \in T_eG$ (for natural gradient) such that for any $X\in T_eG$ we have that
\[
	\omega^{Fisher} (h_{n}, h_{X}) = \mathrm{d}\mathcal{E}\big|_{q_g} [h_X].
\]
By Lemma \ref{lem:EDifferential} and Lemma \ref{lem:multFisher} we have $n = \frac{1}{c_F} \left(\frac{1}{\tau}\E_{q_g}\left[\theta \nabla_{\theta} \ell\right] - \mathbf{1}\right).
$

Thus we know that our direction of descent should be $-n$ and we can absorb $1/c_F$ and temperature $\tau$ into the step-size. Use the exponential map $\exp:T_eG \mapsto G$ followed by $L_g$ as the retraction. Choosing $\alpha c_F \tau  >0$ as a step size, we have
\[
	g^{\text{updated}} = g\exp(-\alpha n) = g \exp\left(- \alpha \left( \E_{q_g}[\theta \nabla_{\theta} \ell] - \tau \mathbf{1} \right)\right).
\]	

Componentwise this reads as
\begin{equation}
	g_i^{\text{updated}} = g_i \exp(- \alpha \left(\E_{q_g}\left[\theta_i \partial_i \ell\right] - \tau\right)).
\end{equation}

This update rule naturally preserves the condition that $g_i>0$, i.e. we stay on the manifold since the exponential map is not only defined locally but on all $T_eG$. This could not be guaranteed in the update rule by \citet{KhRu21} which used an only locally  defined retraction function.

\subsection{Specializing to the diagonal affine group}
\label{app:affine}

\begin{algorithm}[h!]
   \caption{Affine Update}
      \label{fig:affineAlg}
      \definecolor{commentgray}{rgb}{0.25, 0.25, 0.25}
      \begin{algorithmic}[1]
          \REQUIRE $\alpha>0$ step size, $K\geq 1$ MC-sample number, $\beta _1, \beta_2 \in [0, 1)$, $\nu$ an even smooth distribution on $\R$.
          \STATE Put $c_X = \int_\R \Big(1 + \theta \frac{\nu'(\theta)}{\nu(\theta)} \Big)^2 \d \theta $, $c_y = \int_\R \frac{\nu'(\theta)^2}{\nu(\theta)} \d \theta$.
          \STATE Initialize $(A,b) \in \R_{>0}^P\times \R^P$ randomly, and $M_U = M_V = \mathbf{0} \in \R^P$.
          \WHILE{not converged}
          \STATE Sample a minibatch $\mathcal{J} \subseteq [1..N]$ of size $n$,
          \STATE Sample noise vectors $\varepsilon_k \sim \nu^P$ for $k = 1, \ldots, K$,
          \STATE Put $\theta_k = A \varepsilon_k + b$, the MC-parameter samples,
          \STATE $ U_k = \frac{1}{n} \sum_{j \in \mathcal{J}} A\varepsilon_k \nabla_\theta \ell_j(\theta_k) + \frac{1}{N} A\varepsilon_k \nabla_\theta R(\theta_k) - \frac{\tau}{N} \mathbf{1}$,
          \STATE $U \leftarrow \frac{1}{c_X K} \sum_{k =1}^K U_k$,
          \STATE $V_k = \frac{1}{n} \sum_{j \in \mathcal{J}}  A\nabla_\theta \ell_j(\theta_k) + \frac{1}{N}A\nabla_\theta R(\theta_k)$
          \STATE $V \leftarrow \frac{1}{c_y K } \sum_{k = 1}^K V_k$ 
		  \STATE $M_V \leftarrow (1 - \beta_1) V + \beta_1 M_V$,                              
          \STATE $M_U \leftarrow (1 - \beta_2) U + \beta_2 M_U$,          
          \STATE $b \leftarrow b + A\frac{\exp(-\alpha M_U) - I}{M_U} M_V$ 

          \STATE $A \leftarrow A \exp(-\alpha M_U)$,
          \ENDWHILE
          \STATE Return $(A,b)$.
     \end{algorithmic}
     \emph{Loss function: $\ell(\theta) = \sum_{i=1}^N \ell_{i}(\theta) + R(\theta)$. Here, the product of two vectors is again taken to mean componentwise as well as the exponential. The function $(e^{-\alpha x} - 1)/x$ is well defined for all $x$, in evaluating it near $x = 0$ we may simply use the linear Taylor approximation in order to avoid division by $0$.}
\end{algorithm}

\subsubsection{The Affine group and its action}

The affine group combines the freedom of translations of the additive group and the scaling of the multiplicative groups. As with the above two groups we will use the mean field distribution, and therefore the scaling will be componentwise.

Realize the diagonal affine group $\operatorname{Aff}_P^{\diag}(\R)$ as pairs $(A, b)$ where $A$ is a $P\times P$ positive diagonal matrix and $b\in \R^P$. The group operation is given as $(A_1,b_1) (A_2, b_2) = (A_1 A_2, A_1b_2 + b_1).$ The action on parameters $\Theta = \R^P$ is 
\begin{equation}\label{eq:affineAction}
	(A, b) \cdot \theta = A\theta + b.
\end{equation}
This is compatible with the group multiplication defined above meaning $g_1 \cdot (g_2 \cdot \theta) = (g_1g_2) \cdot \theta$, in fact this is why the group operation has been defined in such a way.

The group $\operatorname{Aff}_P^{\diag}(\R)$ can be realized as a subgroup of matrices, $$(A, b) \mapsto \begin{pmatrix} A & b\\ 0 &1 \end{pmatrix} \in \operatorname{GL}_{P+1}(\R)$$ is an injective group homomorphism. This is a Lie group, with a Lie algebra consisting of pairs $\{(X,y): X \text{ a diagonal } P\times P \text{ matrix}, y \in \R^P\}$ and the Lie bracket is given by
\[
	[(X_1,y_1) ,(X_2, y_2) ] = (0, X_1y_2 - X_2 y_1).
\]
The exponential for this group is $\exp_e((X,y)) = (e^X, \frac{e^{X}-1}{X} y)$, as can be seen most easily from the matrix representation of Lie algebra elements as $\left(\begin{smallmatrix} X& y \\ 0&0\end{smallmatrix}\right)$. The meaning of $(e^{X} - I)/X$ is best understood interms of the Taylor expansion, and the expansion begins as $I + \tfrac12 X + \frac{1}{6} X^2 + \cdots$. 

\subsubsection{The information manifold $\mathcal{Q}$ and its geometry}

Pick a base distribution $\tilde{q}_0 \in \mathcal{P}^+(\R)$, 
and put $q_0(\theta) = \prod_{i=1}^P \tilde{q}_0(\theta_i) \in \mathcal{P}^+(\R)$ 
and let $\Qcal =\{q_g(\theta) = \pi(g)q_0(\theta) : g\in G \} = \{\tfrac{1}{|A|} q_0(A\inv (\theta - b)): (A,b) \in \operatorname{Aff}_P^{\diag}(\R)\}.$ Here $|A|$ denotes the absolute value of the determinant of $A$, which is the Jacobian determinant $|\d((A,b)\cdot \theta)/\d\theta)|$.

\begin{lemma}\label{lem:AffineTangents}
Let $q = q_g \in \Qcal$. The following $2P$ tangent vectors form a basis of $T_q\Qcal$:
\[
	h^X_i = - \pi(g) \left(q_0+ \delta_i q_0 \right)  \qquad \text{ and } \qquad h^y_i = -\pi(g) \partial_i q_0. 
\]
for $i = 1, \ldots, P$. Here $(\delta_i f )(\theta) = \theta_i \partial_i f(\theta)$.
\end{lemma}
\begin{proof}
Given $(X,y) \in T_eG$ with $X = \diag((X_1, \ldots, X_P))$ we obtain a tangent vector as $\frac{\d}{\d t} q_{g e^{t(X,y)} }\big|_{t=0}$, call it $h_{(X,y)} = h_{(X,y)}^g $. Calculating explicitly,
\begin{align*}
	h_{(X,y)}(\theta) &= \frac{\d}{\d t} \frac{1}{|A e^{tX}|} q_0\left(e^{-tX}A\inv \theta - e^{-tX}A\inv b + \tfrac{e^{-tX}-I}{tX}ty\right)  \Bigg|_{t = 0}\\
	&= - \sum_i \frac{X_i}{|A|} q_0(A\inv (\theta -b)) 
	- \frac{1}{|A|} \sum_i \partial_i q_0( A\inv (\theta - b)) \left(X_i A_i\inv (\theta_i - b_i) + y_i\right)\\
	&= -\left(\sum_i X_i\right) \pi(g)q_{0}(\theta) - \sum_{i}  X_i \pi(g) \delta_i q_0(\theta) - \sum_i y_i \pi(g) \partial_i q_0(\theta).
\end{align*}
for $g = (A,b) \in G$. Recall $\delta_i q= \theta_i \partial_i q$.  Choosing the standard basis in $(X, y)$ we get the basis with $2P$ elements in the statement of the lemma.
\end{proof}

With respect to this basis calculating the Fisher matrix consists of calculating $\omega_{\text{Fisher}} (h^\bullet_i, h^\bullet_j)$. 

\begin{lemma}\label{lem:AffineGroupFisher}
Given the basis in Lemma \ref{lem:AffineTangents}, the Fisher information matrix is a block diagonal matrix, with $2\times 2$ symmetric blocks of the form $\left(\begin{smallmatrix} A&B\\ B & C\end{smallmatrix} \right)$  corresponding to pairs of basis elements $\{h_i^X, h_i^y\}$ for $i = 1, \ldots, P$.

In general $ B = \int_{\R} \big(1 + \theta\frac{\tilde{q}_0'(\theta)}{\tilde{q}_0(\theta)}\big) \tilde{q}_0'(\theta) \d \theta $.
If $\tilde{q}_0$ is symmetric around the origin as the integrand is an odd function $B = 0$. Other entries are given as
\[
	A = \int_{\R} \left(1 + \theta\frac{\tilde{q}_0'(\theta)}{\tilde{q}_0(\theta)}\right)^2 \tilde{q}_0(\theta) \d \theta \qquad \qquad 
	C = \int_{\R} \frac{\tilde{q}_0'(\theta)^2}{q_0(\theta)}\d \theta.
\]
\end{lemma}

\begin{proof}
	We first show that for $i \neq j$ the vectors are orthogonal with respect to this metric,
	\begin{align*}
		\omega_{\text{Fisher}} (h_i^X, h_j^X) 
		&= \int_{\R^P} \frac{h_i^X(\theta)}{q_g(\theta)} \frac{h_j^X(\theta)}{q_g(\theta)} q_g(\theta) 
		= \int_{\R^P} \frac{\pi(g)(q_0 + \delta_i q_0)(\theta)}{\pi(g) q_0(\theta)} \frac{\pi(g) (q_0 + \delta_j q_0) (\theta)}{\pi(g) q_0(\theta)} \pi(g) q_0(\theta) \d \theta \\
		&= \int_{\R^P} \frac{(q_0 + \delta_i q_0)(\theta)}{q_0(\theta)} \frac{(q_0 + \delta_j q_0) (\theta)}{q_0(\theta)}  q_0(\theta) \d \theta \\
		&= \int_\R \left(1 + \frac{\theta_i \tilde{q}_0'(\theta_i)}{\tilde{q}_0(\theta_i)}\right) \tilde{q}_0(\theta_i)\d \theta_i \int_\R \left(1 + \frac{\theta_j \tilde{q}_0'(\theta_j)}{\tilde{q}_0(\theta_j)}\right)\tilde{q}_0(\theta_j) \d \theta_j \prod_{k \neq i,j} \int_\R \tilde{q}_0(\theta_k) \d \theta_k  .
	\end{align*}  
	The $\theta_k$ integrals are $1$, and the other two integrals both vanish as
	\[
		\int_{\R} \left(1 + \theta \frac{\tilde{q}_0'(\theta)}{\tilde{q}_0(\theta)}\right) \tilde{q}_0(\theta)\d \theta = 1 + \int_\R \theta \tilde{q}_0'(\theta) \d \theta
	\]
	integration by parts on the last integral gives $-1$ hence the integral vanishes.
	
	It should be clear from this calculation that also $\omega_{\text{Fisher}}(h_i^X, h_j^y) =\omega_{\text{Fisher}}(h_i^y, h_j^y)= 0$ for $i \neq j$, and that $\omega_{\text{Fisher}}(h_i^X, h_i^X) = A$, $\omega_{\text{Fisher}}(h_i^X, h_i^y) = B$ and $\omega_{\text{Fisher}}(h_i^y, h_j^y) = C$ for all $i$.  
\end{proof}

Let us now write some special cases of distributions.

For the normal distribution $q_0(\theta) = \frac{1}{\sqrt{2\pi}} e^{-\theta^2/2}$ we have that these blocks are of the form $\left(\begin{smallmatrix} 2&0\\0&1\end{smallmatrix}\right)$. 
For the Cauchy distribution $\tilde{q}_0(\theta) = \frac{1}{\pi}\frac{1}{1 + \theta^2}$ we have the Fisher block $\left( \begin{smallmatrix} 1/2 & 0 \\ 0 & 1/2 \end{smallmatrix}\right)$.

\subsubsection{The Differential of $\mathcal{E}$}

As per \eqref{eq:circled1formula} and \eqref{eq:circled2calculations}, we need to make several calculations in this specific case
\begin{equation}\label{eq:affineexptheta}
	\frac{\d}{\d t} (e^{t(X, y)} \cdot \theta)\bigg|_{t = 0} = \frac{\d}{\d t}\left.\left(e^{tX} \theta + \frac{e^{tX} - \mathbf{1}}{tX} ty\right)\right|_{t =0} = X \theta + y
\end{equation}
and similarly for $g = (A,b)$ 
\begin{equation}\label{eq:Affinegexptheta}
	\operatorname{Ad}_{(A,b)} (X,y) = \frac{\d}{\d t} \left((A,b) e^{t(X, y)} (A\inv, -A\inv b) \right)\bigg|_{t = 0} = (X, Ay - Xb).
\end{equation}
For the second calculation note that
\[
	(A,b) \left(e^{tX}, \frac{e^{tX} - I }{tX}ty \right) \left(A\inv, -A\inv b\right) = \left(e^{tX}, A \frac{e^{tX} - I}{X}y + b - e^{tX} b\right)
\]
making use of the fact that the matrices $A$ and $e^{tX}$ are all diagonal and hence commute. Taking the derivative at $t = 0$ we get the result. Making use of \eqref{eq:affineexptheta} we get that
\begin{equation}\label{eq:AffineAdjointAction}
	\operatorname{Ad}_{(A,b)}(X, y) \cdot \theta = X(\theta - b) + Ay
\end{equation}

\begin{lemma}
	Given $h_{X, y}= \sum_{i} X_i h_i^X + y_i h^y_i$ a tangent vector in $T_{q_g}\Qcal$ the differential is given as $(\mathrm{d}\mathcal{E}(q_{g})) [h_{X, y}] = \operatorname{tr}(N_X^\top X) + n_y^\top y$ where
	\[
		N_X = \frac{1}{\tau} \diag(\E_{q_g} [(\theta - b)\nabla_\theta \ell])- I_{P\times P} \qquad\text{ and } \qquad  n_y = \frac{1}{\tau}A \E_{q_{g}} [\nabla_\theta\ell].
	\]
\end{lemma}

Note that even though we are using the matrix notation for $N_X$ and $X$, together with the Frobenius norm, since both are diagonal matrices, this is simply a dot product of their diagonal vectors.

\begin{proof}
	We continue from \eqref{eq:circled1formula}, \eqref{eq:circled2calculations}. Due to \eqref{eq:affineexptheta} and \eqref{eq:AffineAdjointAction} we have that
	\begin{align*}
		\circled{1} &= \int q_g(\theta) (\nabla_\theta \ell(\theta))^\top (X(\theta - b) + Ay) \d \theta \\
		&=\sum_i X_i \int q_g(\theta) (\theta_i - b_i)\partial_i\ell(\theta) \d \theta + \sum_i y_i A_i\int q_g(\theta) \partial_i \ell (\theta) \d \theta  
	\end{align*}
	and
	\begin{align*}
		\circled{2} &= \int (\nabla_\theta q_0(\theta))^\top (X \theta + y) \d\theta \\
		&= \sum_i X_i \int \theta_i \partial_i q_0(\theta) \d \theta + \sum_i y_i \int \partial_i q_0(\theta) \d \theta\\
		&= \sum_i X_i \cancelto{-1}{\int_{-\infty}^\infty \theta_i \tilde{q}'_0(\theta_i)}  \d \theta_i + \sum_i y_i \cancelto{0}{\int_{-\infty}^\infty \tilde{q}'_0(\theta_i) \d} \theta_i.
	\end{align*}
	We get the differentials above when we combine these two calculations. 
\end{proof}

\subsubsection{The Update Rule}

We combine the calculations above assuming $\tilde{q}_0$ is an even function and hence the Fisher matrix is simply diagonal, we write it as  $\diag(c_X, c_y)$. This means that  
\[
	(A^\text{updated}, b^{\text{updated}}) = (A,b) \exp\left(-\alpha (\tfrac{1}{c_X}n_x, \tfrac{1}{c_y}n_y)\right) 
\]

Now let $\alpha \mapsto \alpha \tau c_X$ and in separate coordinates this reads as
\begin{align}
	A_i^\text{updated} &= A_i \exp\left(-\alpha (\E_{q_{g}}[(\theta_i - b_i) \partial_i \ell] - \tau)\right) \label{eq:AppDiagAffineAUpdate}\\
	b_i^{\text{updated}} &= b_i + \frac{c_X}{c_y}A_i^2 \frac{\exp\left(-\alpha(\E_{q_g}[(\theta_i - b_i)\partial_i\ell] - \tau)\right)-1}{\E_{q_g}[(\theta_i - b_i)\partial_i\ell]-\tau} \E_{q_g}[\partial_i \ell].\label{eq:AppDiagAffineBUpdate}
\end{align}

Notice that the condition $A_i > 0$ is preserved by this update rule. In BLR the positive definiteness of the covariance matrix parameter of a multivariable gaussian distribution cannot be preserved with linear updates, save for very small step sizes. This issue was later remedied by another method by \citet{LiSc20}, where the authors used a quadratic approximation to the geodesic on the manifold $\Qcal$ which also satisfied the positive definiteness constraint.

As a special case consider when $q_0$ is the Dirac delta distribution at $0\in \R^P$. Our derivation does not work, but the above update rules are still valid.

The expectation $E_{q_g}[(\theta - b)\nabla_\theta \ell] = 0$ in the Dirac delta case, which makes the updates on $A$ components a moot point. This is to be expected since reducing to the Dirac-delta case means willfully forgoing any variance consideration. As for the $b_i = \theta_i$ update the ratio of exponentials only scales our step size and we have
\[
	\theta^{\text{updated}} = \theta - \alpha' \nabla_\theta \ell(\theta)
\]
where $\alpha' = \alpha \tfrac{c_X}{c_y} \frac{e^{\alpha T} - 1}{\alpha T}$ is the modified step size. In other words we simply get the usual gradient descent update rule.

\subsection{Linear approximation gives the BLR}
\label{app:KhanRue}

\subsubsection{The multiplicative case}

The Bayesian Learning Rule (BLR) of \citet{KhRu21} is given for exponential families, but our scheme also includes exponential families such as the family of exponential distributions $\tilde{q}_0 = e^{-\theta}$ as mentioned in the beginning of this section. Then $q_{1/\lambda}(\theta) = (\prod_i\lambda_i) e^{-\langle \theta,\lambda\rangle }$ where $1/\lambda$ is interpreted componentwise.

We write the rule for $\lambda$'s noting that $g_i\inv = \lambda_i$. Then using the linearization of exponential
\begin{equation}\label{eq:multApproxIsBLR}
	\lambda_i^{\text{updated}} = \lambda_i \exp\left( \alpha(\E_{q_{1/\lambda}} [\theta_i \partial_i \ell(\theta)] - 1)\right) \approx  \lambda_i ( 1 - \alpha) + \alpha \lambda_i \E_{q_{1/\lambda}}[\delta_i \ell]
\end{equation}
where we took the linear approximation of the exponential. Note that the Fisher matrix $F_{q_{1/\lambda}} = \diag(\lambda_i^{-2})$ and that $$\nabla_\lambda \E_{q_{1/\lambda}}[\ell] = \diag\left(\tfrac{1}{\lambda_i} \E_{q_{1/\lambda}}[\delta_i \ell]\right).$$ The right hand side of \eqref{eq:multApproxIsBLR} is exactly the rule of \citet{KhRu21} applied to the family of exponential distributions.

\subsubsection{The affine case}
Again we expect that the linear approximation to the above update rule to give us the update rule from \citet{KhRu21} when we are in the case of the diagonal Gaussian distributions. This happens under temperature $\tau = 1$. 

If $q_0(\theta) = \frac{1}{\sqrt{2\pi}^P} e^{-\frac12 \theta^\top I \theta}$ then we see that the family is the space of diagonal-covariance Gaussian distributions. 
\[
	q_{(A,b)}(\theta) = \frac{1}{\sqrt{2\pi}^P|A|} e^{-\frac{1}{2} (\theta - b)^\top (A^2)^{-1} (\theta - b)}
\]
therefore we see that in the notation of Normal distributions $\mathcal{N}(m, \Sigma)$ we have $A^2 = \Sigma = {S}\inv$ (all diagonal matrices in our case) and $b = m$. 

For $b$ the update rule has a linear approximation in $\alpha$,
\[
	b^{\text{updated}} \approx b - 2\alpha A^2 \E_{q_{g}}[\nabla_\theta \ell]
\]	
where we used that $\frac{e^{-\alpha X }- 1}{X} = -\alpha + \alpha^2X/2 + \cdots$ and that for our chosen $\tilde{q}_0$ that $c_X/c_y = 2$. 

This agrees with the first part of \citet[Eq.~12]{KhRu21} save for the factor of $2$.

The second part of the same equation is written in terms of an ${S}$ update, so we take \eqref{eq:AppDiagAffineAUpdate} and turn it into an update about ${S}$ by inverting and squaring both sides:
$
	{S}^{\text{updated}}_i = {S}_i \exp\left(2\alpha(\E_{q_g}[(\theta_i - b_i) \partial_i \ell] - 1)\right)$
	
Taking the linear approximation in $\alpha$,
\[
	{S}^{\text{updated}}_i \approx {S}_i(1- 2\alpha) + 2\alpha \E_{q_{g}}[(\theta_i- b_i) \partial_i \ell] 
\]
where ${S}$ may be considered as either a vector or a diagonal matrix. The right hand side may also be written as
\[
 {S}^{\text{updated}} \approx {S}(1- 2\alpha) + 2\alpha \E_{q_{g}}[\nabla_\theta^2\ell]
\] 
following an application of integration by parts using the special form of the Gaussian measure $q_{g}$. This is exactly the second update rule in \citet[Eq.~12]{KhRu21} (with $\alpha\mapsto \alpha/2$).

\section{DETAILS OF THE EXPERIMENTS}
For MNIST, no data augmentation was considered. All CIFAR and TinyImageNet experiments use basic data augmentations (random horizontal flipping and cropping). To account for the data augmentation,
we set $N \leftarrow 4 \cdot N$. All hyper parameters were selected via a grid search over a moderate amount of configurations and we selected the ones giving the best results for each method.

\subsection{Additive vs. multiplicative learning}
\label{app:experiment1}
In all experiments, the learning rate is annealed to zero using a cosine scheduler, and we used $5$ ``warm-up'' epochs where $\alpha$ is linearly increased from zero to the starting learning rate.

\paragraph{MLP.} We used a fully connected network with 5 hidden layers (1024, 512, 256, 256, 256 neurons) and $\tanh$ nonlinearity.
The regularizer was fixed to $R(\theta) = 125 \cdot \| \theta \|^2$. We train for $60$ epochs.
The additive update~(\Cref{fig:additiveAlg}) runs with $\alpha = 0.01$, $\beta = 0.9$, $q_0 = \mathcal{N}(0, 0.001)$, 
For the multiplicative update~(\Cref{fig:multAlg}) we used $\alpha = 50$, $\beta = 0.9$, set $q_0$ as a Rayleigh distribution (with parameter set to $1$), and fixed $\tau = 0.005$.
Both methods use $K=32$ MC samples and a batch size of $n=50$.

\paragraph{CNN.} The convolutional neural network is a basic LeNet-5 architecture with 128, 256 and 512 convolutional filters in the layers. Each convolution is followed by a max-pooling and we used two fully connected layers with 512 and 256 neurons at the end of the network. The regularizer was fixed to $R(\theta) = \| \theta \|^2$. We train for $180$ epochs. 
The additive and multiplicative updates were run using the same parameters as described in the MLP paragraph, except that we used $\alpha = 100$ and $\tau = 0.001$ for the multiplicative updates.
For the CNN experiments, both methods use $K=10$ MC samples and batch size $n = 100$.

\begin{table}[h!]
  \centering
\begin{tabular}{l|llll}
  Algorithm & $\alpha$ & $\beta_1$ & $\beta_2$ & damping \\
  \midrule 
  Affine rule~(\Cref{fig:affineAlg}) & 1 & 0.8 & 0.999 & -- \\
  Additive rule~(\Cref{fig:additiveAlg}) & 0.1 & 0.8 & -- & -- \\
  iVON~\citep[Figure~1]{LiSc20} & 0.5 & 0.8 & 0.999 & 1 \\
  VOGN~\citep{OsSw19} & 0.002 & 0.95 & 0.999 & 0.01 \\
  SGD & 0.1 & 0.8 & -- & -- 
\end{tabular}
\caption{Hyperparameters for the experiments shown in~\Cref{table:additive_vs_affine}.}
\label{tab:hyper}
\end{table}

\subsection{The affine learning rule}
\label{app:experiment2}
In all experiments, we use the ResNet-20 architecture (as in \citet[Table~6]{HeZh16}) with filter response normalization and train for $180$ epochs. The learning rate is decayed to zero using a cosine learning rate scheduler. All methods use one MC sample ($K = 1$). The regularizer is fixed to $R(\theta) = 50 \cdot \| \theta \|^2$.  

The hyperparameters for the individual methods are fixed across the data sets and given in~\Cref{tab:hyper}.

The three different choices of $q_0$ are: $\theta \sim \mathcal{N}(0, 25 \cdot 10^{-6})$, $\theta \sim 0.005 \cdot \text{Laplace}(0, 1)$, and $\theta \sim \text{Uniform}[-0.0025, 0.0025]$.

\end{document}